\pdfoutput=1

\documentclass[sigconf,authorversion=true]{aamas} 

\usepackage[utf8]{inputenc}
\usepackage[active]{srcltx}
\usepackage[english]{babel}
\usepackage{verbatim}
\usepackage{mathtools}
\usepackage{bm}
\usepackage{bbm}
\usepackage{amsmath}
\usepackage{microtype,comment}
\usepackage{booktabs} 
\usepackage{hyperref}
\usepackage{verbatim}
\usepackage{mathtools}
\usepackage{bm}
\usepackage{amsmath} 
\usepackage{amsthm}
\usepackage{dsfont}
\usepackage{array}
\usepackage{mathrsfs}
\usepackage{color}
\usepackage{natbib}
\usepackage{enumerate}
\usepackage[ruled]{algorithm2e}
\usepackage{algpseudocode}
\usepackage[font=footnotesize, skip=3pt]{subcaption}
\usepackage[font=footnotesize, skip=3pt]{caption}

 \bibpunct[, ]{(}{)}{,}{a}{}{,}%
 \def\newblock{\ }%
\theoremstyle{plain} 
\newtheorem{lemma}{\textbf{Lemma}} 
\newtheorem{proposition}{\textbf{Proposition}}
\newtheorem{theorem}{\textbf{Theorem}}
 
\newtheorem{assumption}{\textbf{Assumption}}

\theoremstyle{definition}
\theoremstyle{remark}

\DeclareMathOperator*{\argmax}{arg\,max}

\newcommand{\betadt}{\beta\Delta t}
\newcommand{\gammadt}{\gamma\Delta t}

\newcommand{\E}{\mathbb{E}}

\usepackage{balance} % for balancing columns on the final page

\setcopyright{ifaamas}
\acmConference[AAMAS '23]{Proc.\@ of the 22nd International Conference
on Autonomous Agents and Multiagent Systems (AAMAS 2023)}{May 29 -- June 2, 2023}
{London, United Kingdom}{A.~Ricci, W.~Yeoh, N.~Agmon, B.~An (eds.)}
\copyrightyear{2023}
\acmYear{2023}
\acmDOI{}
\acmPrice{}
\acmISBN{}

\acmSubmissionID{674}

\title{Equilibrium Bandits: Learning Optimal Equilibria of Unknown Dynamics}
\author{Siddharth Chandak}
\affiliation{
  \institution{Stanford University, Department of Electrical Engineering} 
  \country{USA}}
\email{chandaks@stanford.edu}

\author{Ilai Bistritz}
\affiliation{
  \institution{Stanford University, Department of Electrical Engineering} 
  \country{USA}}
\email{bistritz@stanford.edu}

\author{Nicholas Bambos}
\affiliation{
  \institution{Stanford University, Department of Electrical Engineering}
  \country{USA}}
\email{bambos@stanford.edu}

\begin{abstract}
Consider a decision-maker that can pick one out of $K$ actions to control an unknown system, for $T$ turns. The actions are interpreted as different configurations or policies. Holding the same action fixed, the system asymptotically converges to a unique equilibrium, as a function of this action. The dynamics of the system are unknown to the decision-maker, which can only observe a noisy reward at the end of every turn. The decision-maker wants to maximize its accumulated reward over the $T$ turns. Learning what equilibria are better results in higher rewards, but waiting for the system to converge to equilibrium costs valuable time. Existing bandit algorithms, either stochastic or adversarial, achieve linear (trivial) regret for this problem. We present a novel algorithm, termed Upper Equilibrium Concentration Bound (UECB), that knows to switch an action quickly if it is not worth it to wait until the equilibrium is reached. This is enabled by employing `convergence bounds' to determine how far the system is from equilibrium. We prove that UECB achieves a regret of  $\mathcal{O}(\log(T)+\tau_c\log(\tau_c)+\tau_c\log\log(T))$ for this ``equilibrium bandit problem'' where $\tau_c$ is the worst case approximate convergence time to equilibrium. We then show that both epidemic control and game control are special cases of equilibrium bandits, where $\tau_c\log \tau_c$ typically dominates the regret. We then test UECB numerically for both of these applications.

\end{abstract}

\medskip

\begin{document}

%%% The following commands remove the headers in your paper. For final 
%%% papers, these will be inserted during the pagination process.

\pagestyle{fancy}
\fancyhead{}

%%% The next command prints the information defined in the preamble.

\maketitle 

\section{Introduction}\label{sec-intro}

Many large-scale complex systems reach an equilibrium over time. Examples include epidemics, transportation, markets, and supply chains. With no planning, the global performance at this equilibrium can be poor. When a decision-maker can control some parameters of such a system, it can influence the equilibrium that the system converges to. Examples are changing the frequencies of subway lines, or the masking and isolation policies during an epidemic. However, a model for the dynamics of these large-scale complex systems is rarely available. Instead, the decision-maker can only observe the impact of its decisions in real time. It is infeasible to allow the system to converge to equilibrium under each policy since this will waste significant time on suboptimal policies. This introduces a learning problem of controlling such systems with `bandit feedback' \cite{bandit-book}. 

Motivated by this, we consider an agent that takes an action from a discrete set of actions at each timestep. There is an underlying system that evolves with time depending on the agent's action. At each timestep, the agent receives a noisy reward as a function of its action \textbf{and} the  `state' of the system. The key aspect of the underlying system is that if we fix the action, it would asymptotically converge to a unique equilibrium as a function of this action. However, waiting too long for the system to converge to a bad equilibrium is costly. We measure the regret of the agent as the difference between the reward at the optimal equilibrium (i.e., for the optimal action) and the accumulated reward of the agent. This introduces a new bandit problem which we name \textit{`Equilibrium Bandits'}.

The reward process in equilibrium bandits is not i.i.d. over time. The expected reward approaches the expected reward at equilibrium and therefore has memory. Hence, stochastic bandit algorithms result in linear regret in $T$. On the other hand, adversarial bandit algorithms also result in linear regret. 

In applications such as epidemics and transportation, convergence to equilibrium can take significant time. Therefore, we are interested in the dependence of the regret on the worst-case `approximate convergence time' $\tau_c$ in addition to the horizon $T$. Alternatively, we can think of $\tau_c$ as $T$-dependent. 

When estimating the reward at equilibrium, the distance of the current state from equilibrium creates `equilibrium noise'. Hence, we need more consecutive `arm pulls' to weaken the equilibrium noise at the last pull. This conflicts with the averaging required to weaken the i.i.d. reward noise, since averaging over longer periods would have to use states that are too far from equilibrium. 

We present the Upper Equilibrium Concentration Bound (UECB) algorithm for equilibrium bandits that builds upon the basic intuition behind the Upper Confidence Bound (UCB) algorithm \cite{bandit-book}. Our main innovation is employing `convergence bounds' to determine the maximum possible reward the agent could get by waiting for the system to converge to equilibrium for a given action. A chosen action in UECB is played consecutively for a full ``epoch''. The epoch length increases with the number of epochs this action has been chosen in the past. Consequently, UECB spends time weakening the equilibrium noise only for promising actions. UECB balances between the i.i.d.\ noise and equilibrium noise by only using a fraction of samples from a given epoch for the reward estimation. We show that UECB achieves $\mathcal{O}(\log(T)+\tau_c\log(\tau_c)+\tau_c\log\log(T))$ expected regret. We also prove a lower bound that shows that UECB is optimal up to logarithmic factors in $\tau_c$.

We detail two real-life problems that can be modeled as equilibrium bandits. The first example is epidemic control, where the agent is the government that chooses a policy that may include lockdowns, link closures, and enforcing masks \cite{bistritz2019controlling}. These policies incur both an operational cost and a health cost by affecting the infection rate. We use the SIS model \cite{SIS1} to formalize the epidemic spread. Our second example is a continuous game where the system consists of multiple players trying to optimize their utility functions using gradient-based learning \cite{mazu}. Specifically, we consider resource allocation games where the policymaker chooses the set of resources available to each player and wishes to maximize the sum of utilities of all players at equilibrium \cite{korilis1999avoiding}. We then simulate these examples and show that UECB performs well based on minimal knowledge of the system that is readily available in practice. 

\subsection{Related Work}
There has been plenty of research on multi-armed bandits in the last century, including stochastic bandits, contextual bandits, linear bandits, and adversarial bandits \cite{bandit-book}. Special cases for each of these have been studied in great detail \cite{bandit-1, bandit-2, Auer_adversarial}. There are also works on bandits which deal with a system evolving with time. These often deal with a Markov-chain based stochastic evolution, e.g., restless bandits \cite{Whittle_1, Whittle_2}. 

Unlike restless bandits, in equilibrium bandits, an unperturbed system, where the chosen action is fixed, would asymptotically converge to equilibrium as a function of that action. Such behavior is typical to multiagent systems, specifically with many humans in the loop (e.g., epidemics and transportation).  Furthermore, stochastic and adversarial bandit algorithms both give linear regret in equilibrium bandits. We propose the first algorithm, named UECB, that achieves sublinear regret for equilibrium bandits. 

Equilibrium bandits can be thought of as a special case of non-stationary stochastic bandits \cite{besbes2014stochastic}, but applying this approach would result in a regret bound of $O(T^\frac{2}{3})$ at best (depending on the convergence rate to equilibrium) since it does not leverage the converging structure of equilibrium bandits. 

Another field closely related to our work is that of reinforcement learning (RL) \cite{Sutton-book}. Classically, the RL problem is modeled using Markov decision processes and the aim of the agent is to choose the action at each step that maximizes its cumulative reward. There has been significant research in developing algorithms for RL \cite{Watkins, BVR, policy_grad}. In recent years, there has been progress in developing deep learning-based methods for complex problems such as multi-agent control, robotics, and games\cite{alphago, atari, dec_trans, robotics}. Equilibrium bandits differ from the RL literature in one major aspect: the state in our case evolves in a non-stochastic, converging way. One could model our evolution as a deterministic Markov chain. However, RL algorithms are designed for general problems and typically assume an ergodic Markov chain \cite{Watkins} or the existence of an offline simulator that allows `restarts' of the Markov chain \cite{alphago}. These assumptions do not hold in our case, since our deterministic Markov chain is absorbing and our target is real-time learning which cannot be restarted. Furthermore, in equilibrium bandits, only the reward is observable whereas typical RL assumes that the state is observable as well. 

Our work is related to the literature on control and intervention in games, where a manager can tune some parameters in the reward functions of the players \cite{grammatico2017dynamic,parise2020analysis,mguni2019coordinating, Alpcan2009a,bistritz2021online,ratliff2020adaptive}. While our dynamics do not have to stem from a game, games are a key example of a system that converges to equilibrium. From this point of view, our work is the first to provide regret guarantees while learning to control an unknown game. 

\section{Problem Formulation}\label{sec-prob_form}

Consider an agent that chooses an action $a_t$ at each time $t$ from the action set $\mathcal{A}=\{1,\ldots,K\}$. The action controls an underlying system that evolves with time and affects the agent's reward. Let $z_t$ be the state of the system at time $t$. We assume that $z_t$ lies in a bounded and closed set $\mathcal{Z}\subset\mathbb{R}^d$. Then we define the `evolution' function $g:\mathcal{A}\times\mathcal{Z}\rightarrow\mathcal{Z}$ and the `reward' function $f:\mathcal{A}\times\mathcal{Z}\rightarrow\mathbb{R}$. We assume that $f(a,z)$ is Lipschitz continuous with parameter $L>0$, as a function of $z$, for all actions $a$. Without loss of generality, we make the following two assumptions to simplify the notation:
$f(a,z)$ is bounded in $[0,1]$ for all $a,z$ and $\mathcal{Z}\subseteq \mathcal{B}_{0.5}(\|\cdot\|)$ where $\mathcal{B}_{0.5}(\|\cdot\|)=\{z\in\mathbb{R}^d\mid\|z\|\leq 0.5\}$. Here $\|\cdot\|$ denotes any compatible norm on $\mathbb{R}^d$. The function $g$ determines the next state of the system based on the current state and the action taken, i.e., $z_{t+1}=g(a_t;z_t)$. The function $f$ determines the agent's expected reward, i.e., $x_t=f(a_t;z_t)$. The noisy reward $y_t$ observed by the agent is given by $y_t= x_t+\eta_t$ where $\eta_t$ is i.i.d.\ subgaussian noise with parameter $\sigma$, i.e., $\E[\eta_t]=0$ and $\E[\exp(\alpha\eta_t)]\leq \exp(\alpha^2\sigma^2/2)$ for all $\alpha\in\mathbb{R}$.  Rewards are typically noisy since the effectiveness of a policy cannot be deduced accurately and is often based on stochastic data.

Motivated by applications such as epidemic control and game control, we make the following assumption on the evolution function:
\begin{assumption}\label{assu-evolution}
The function $g(\cdot;\cdot)$ satisfies the following conditions:
\begin{enumerate}[(a)]
    \item For each action $a$, consider the iteration $z_{t+1}=g(a;z_t)$ for $t>0$. There exists a unique equilibrium corresponding to action $a$, i.e., a $z_a^*$ such that $z_a^*=g(a;z_a^*)$. Furthermore, this equilibrium is a stable point, i.e., $\lim_{t\uparrow\infty} g^{(t)}(a;z)=z_a^*$ for all $z\in\mathcal{Z}$.
    
    \item When action $a$ is played, the distance of the state $z_t$ from $z_a^*$ decreases. Formally,
    \begin{equation}\label{assumption-2}
        \|g(a;z)-z_a^*\|\leq c(a;z)\|z-z_a^*\|, \; \forall \; a\in\mathcal{A}, z\in\mathcal{Z},
    \end{equation}
    where $0<c(a;z)<1$. We assume that $c(a;z)$ is bounded away from $1$, i.e., $\exists \; \tau_c\geq 1$ s.t.\ $c(a;z)<e^{-\frac{1}{\tau_c}}$ for all $a$ and $z$. 
\end{enumerate}
\end{assumption}

Part (a) of the above assumption implies that if the agent keeps the action $a$ fixed, the system will asymptotically converge to the equilibrium state $z_a^*$. In addition, we define $x_a^*\coloneqq f(a,z_a^*)$ as the equilibrium reward for action $a$. 

The $\tau_c$ in part (b) is the ``approximate convergence time to equilibrium'', i.e., the timesteps required until the distance of the state from equilibrium is at most a $1/e$ factor of its initial value.

A basic class of functions $g$ satisfying the above assumption are contraction mappings. A contraction mapping has a unique fixed point which is the unique equilibrium point required for part (a). Similarly, the contraction factor is the constant $c(a;z)$. Contraction mappings are commonly found in solutions of many ODEs such as Newton's method \cite{Protter} and for popular policy evaluation schemes such as temporal difference learning \cite{TD0}. In section \ref{sec-appli}, we detail two applications that result in non-contractive mappings but still satisfy the above assumption.

We now return to the problem and the objective of the agent. The agent takes an action $a_t$ at each time $t$. The agent does not observe the underlying state $z_t$ and only observes a noisy version of the reward $f(a_t;z_t)$ at each time $t$. The optimal action $a^*$ is defined as $a^*\coloneqq\argmax_a f(a;z_a^*)$, i.e., the action with the highest equilibrium reward. For simplicity, we assume that this optimal action is unique, i.e., $f(a,z^*_a)<f(a^*,z^*_{a^*})$ for all $a\neq a^*$. Nevertheless, our analysis follows with minor modifications for the case of multiple optimal actions, using the same algorithm. Define the suboptimality gap for each action as $\Delta_a=x^*_{a^*}-x^*_a$. The regret till time $T$ is $R(T)\coloneqq \sum_{t=1}^T \left(f(a^*,z_{a^*}^*) - y_t\right)$. 
We want to design an algorithm for the agent that minimizes the expected cumulative regret:
$$\E[R(T)]=\max_a\sum_{t=1}^T f(a;z^*_a)-\mathbb{E}\left[\sum_{t=1}^Tf(a_t;z_t)\right]$$
where the expectation is with respect to the stochastic noise. With multiple equilibria, we cannot guarantee to which one the system would converge. We can then redefine the above regret such that $z^{*}_{a^{*}}$ is the worst equilibrium for action $a$, with no modifications needed in our analysis or algorithm. 

With this regret, a good algorithm would find the optimal action as quickly as possible and then commit to it to allow the system to converge to the corresponding equilibrium. This objective is inspired by applications such as epidemics and transportation (studied in detail in section \ref{sec-appli}). In an epidemic, the government is the agent that has to choose the best policy to control the spread. Then, the different policies (e.g., lockdown, masks) are the actions and the underlying state is the fraction of infected individuals. The cost then takes into account the health costs (e.g., deaths and complications) and the operational cost (e.g., treatment and economic implications). This example makes it clear why we wish to maximize the expected cumulative reward and why we wish to commit to the optimal action as quickly as possible. 

Our contribution is the novel UECB algorithm, which is presented in the next section. Our main result proves a regret bound for UECB, which we show is optimal up to logarithmic factors. 
\begin{theorem}\label{thm-noisy}
For an equilibrium bandit instance satisfying Assumption \ref{assu-evolution}, with a Lipschitz continuous reward function bounded in $[0,1]$ and noise $\eta_t$ that is subgaussian with parameter $\sigma$, the expected cumulative regret of the UECB algorithm (Algorithm \ref{algo-noisy}) is bounded as:
\begin{eqnarray}\label{eqn-thm-noisy}
    \E[R(T)]&=& \mathcal{O}\bigg(\sum_{a\in\mathcal{A},a\neq a^*}\frac{\sigma^2\log(T)}{\Delta_a}+\tau_c\log\left(\tau_c\log\left(\frac{1}{\Delta_a}\right)\right)\nonumber\\
    &&\;\;\;\;\;\;+\tau_c\log\left(\frac{\sigma^2\log(T)}{\Delta_a^2}\right)\bigg),
\end{eqnarray}
where $\Delta_a=x_{a^*}^*-x_a^*$ is the suboptimality gap for action $a$.
\end{theorem}

The dominant term in \eqref{eqn-thm-noisy} depends on $\frac{T}{\tau_c}$ which quantifies how many times we can afford to converge to equilibrium, and is application dependent. For example, $\E[R(T)]=\mathcal{O}(\sqrt{T}\log(T))$ for $\tau_c=\mathcal{O}(\sqrt{T})$, and $\E[R(T)]=\mathcal{O}(\log(T)\log\log(T))$ for $\tau_c=\mathcal{O}(\log(T))$.

Stochastic bandits can be viewed as a special case of equilibrium bandits where the convergence to equilibrium is instantaneous. Equilibrium bandits is a more challenging problem since the rewards are no longer i.i.d. over time. Algorithms like UCB treat the rewards as independent over time and do not account for how far the system is from the equilibrium. Consequently, the UCB algorithm achieves linear regret for equilibrium bandits (Theorem \ref{UCB-linear} in the Appendix). 

Equilibrium bandits are a special case of adversarial bandits, where any sequence of rewards is allowed. However, adversarial regret bounds are significantly weaker than our regret bound since they compare to the best action in hindsight. In contrast, our regret resembles the more demanding regret of stochastic bandits, which compares to the ``absolute'' optimal action. Therefore, adversarial bandit algorithms also achieve linear regret for equilibrium bandits. 

In our notation, the regret for the adversarial problem would be $$\E[R_{adv}(T)]=\max_a\sum_{t=1}^T f(a;z_t)-\mathbb{E}\left[\sum_{t=1}^Tf(a_t;z_t)\right].$$ 
The adversarial regret looks at the state sequence $\{z_t\}_{t=1}^T$ as given and ignores the fact that our action sequence $\{a_t\}_{t=1}^T$ impacted the state sequence. In the epidemic control example, this would mean that the adversarial regret tries to find the best action given the number of infections over time, as though these numbers could not be avoided by an agent who would have taken better actions. 

\section{UECB Algorithm}
In this section, we present our novel UECB algorithm designed for equilibrium bandits. To provide intuition, we start by analyzing the simpler case where the rewards are not noisy. 

\subsection{The Noiseless Case}\label{subsec:noiseless}
Consider the special case where the rewards are not noisy, i.e., $\eta_t=0$ a.s. for all $t$. Hence, the agent directly observes $y_t=x_t=f(a_t;z_t)$. Since there is no noise, it should be possible to find the optimal action in bounded time which results in an expected regret of $\mathcal{O}(1)$.

A naive algorithm would pick each action consecutively a sufficiently large number of times (denoted by $t_{try}$), to allow the system to approach the equilibrium corresponding to this action. The agent will then know the reward at equilibrium for each action with arbitrarily low error, and can then commit to the optimal action. This naive algorithm is actually the default choice in many real-life scenarios. This algorithm can achieve sublinear regret if $t_{try}$ is above a threshold, which depends on $\tau_c$ and the suboptimality gap. Since the suboptimality gap is unknown to the agent, the naive algorithm achieves linear regret in general.

Instead, we propose the ``Upper Equilibrium Confidence Bound'' (UECB) algorithm for the noiseless case based on `convergence bounds'. Suppose the state of the system at time $t$ is $z_t$ and action $a$ is taken for $\ell$ timesteps consecutively after that. Then, using Assumption \ref{assu-evolution}, we have $$\|z_{t+\ell}-z_a^*\|\leq e^{-\frac{\ell}{\tau_c}}\|z_{t}-z^*\|.$$ Using the assumption that $\mathcal{Z}\subseteq \mathcal{B}_{0.5}(\|\cdot\|)$, we have $\|z_{t}-z^*\|\leq 1$. Without this assumption, the only modification required would be to replace $L$ with $2L\max_{z\in\mathcal{Z}}\|z\|$ henceforth in the paper. Now using the Lipschitz property of the reward function, we deduce that
\begin{equation}\label{UECB-intuition}
    f(a;z_{t+\ell})-Le^{-\frac{\ell}{\tau_c}}\leq f(a;z_a^*)\leq f(a;z_{t+\ell})+Le^{-\frac{\ell}{\tau_c}}.
\end{equation}

We assume that the agent knows $\tau_c$, which is a worst-case bound on the actual convergence time. Given this knowledge, $f(a;z_{t+\ell})+Le^{-\ell/\tau_c}$ is the maximum possible reward action $a$ can yield at the equilibrium point corresponding to $a$. As we demonstrate in the applications of Section \ref{sec-appli}, knowing a bound on $\tau_c$ is significantly easier than knowing the system parameters. 

In practice, the agent can often observe more than just the reward (e.g., the state $z_t$ or the convergence rate $c(a;z_t)$). With more knowledge, the above bounds can be tightened without affecting our analysis or the UECB algorithm that uses them as input. 

It is necessary to play an action consecutively for some time to get an accurate enough estimation of the reward at equilibrium given that action. However, we do not want to always wait for the system to converge as it might waste precious time and incur significant regret. The idea behind the UECB algorithm (Algorithm \ref{algo-noisy}) is to play actions that seem to lead to good equilibria for an increasing number of turns to allow it to reach closer to convergence. This along with the above-mentioned convergence bound serves as the basic intuition behind UECB. Instead of switching an action at every timestep, UECB chooses an action to be played consecutively over a full `epoch'. The epoch length increases with the number of epochs the chosen action has been played before. 

 Let $m_{a,n}$ denote the number of epochs action $a$ has been chosen for till the end of epoch $n$, and let $t_n$ denote the total number of timesteps till the end of epoch $n$. Additionally, let $t_{a,n}$ denote the number of timesteps action $a$ has been played till the end of epoch $n$. Let the action in the next epoch $a_{n+1}$ be chosen as $$a_{n+1}=\argmax_{a\in\mathcal{A}}UECB_{a,n}$$ where $UECB_{a,n}$ is defined below. The length of epoch $n+1$ is chosen as $\ell_{n+1}\coloneqq2\rho_2\exp(\rho_1(m_{a_{n+1},n}+1))$, where $\rho_1,\rho_2$ are positive parameters, as explained after Theorem \ref{thm-noiseless}. The agent plays this action for the complete epoch and observes the reward obtained at the last timestep of that epoch $x_{t_{n}+\ell_{n+1}}=x_{t_{n+1}}$, denoted by $\hat{x}_{a,n+1}$. Finally at end of epoch $n+1$, $UECB_{a_{n+1},n+1}$ is updated as follows:
$$UECB_{a_{n+1},n+1}=\hat{x}_{a,n+1}+Le^{-\frac{\ell_{n+1}}{\tau_c}}$$
and $UECB_{a,n+1}=UECB_{a,n}$ for all other actions $a\neq a_{n+1}$.

We now give a bound on the maximum number of times the UECB algorithm chooses a non-optimal action and a bound on the maximum possible regret in the noiseless case.
\begin{theorem}\label{thm-noiseless}
Let $\eta_t=0, a.s.$ for all $t$ (i.e., no noise). Then for an equilibrium bandit instance satisfying Assumption \ref{assu-evolution}, with a Lipschitz continuous reward function bounded in $[0,1]$, 
\begin{enumerate}[(a)]
    \item Algorithm \ref{algo-noisy} chooses a suboptimal action, i.e., action other than $a^*$, only for a finite number of turns $\hat{T}$ where $$\hat{T}= \mathcal{O}\left(\sum_{a\in\mathcal{A},a\neq a^*}\tau_c\log\left(\frac{1}{\Delta_a}\right)\right).$$
    \item For all $t>0$,
$$    R(t)= \mathcal{O}\left(\sum_{a\in\mathcal{A},a\neq a^*}\Delta_a\tau_c\log\left(\frac{1}{\Delta_a}\right) + \tau_c\log\left(\tau_c\log\left(\frac{1}{\Delta_a}\right)\right) \right).$$
\end{enumerate}
\end{theorem}

Part (a) of Theorem \ref{thm-noiseless} is based on the maximum number of times a suboptimal action may need to be played consecutively to differentiate it from the optimal action. This gives the maximum number of epochs that may be required for each suboptimal action and hence the number of steps required in the worst case. The first term in the regret bound is obtained by simply multiplying the suboptimality gap for each action. The second term stems from the maximum number of times UECB switches between actions. Switching to a new action resets the convergence of the system to a new equilibrium,  which incurs $\mathcal{O}(\tau_c)$ regret per switch.

We can construct scenarios where any algorithm that achieves sublinear regret would have to play each suboptimal action $a$ at least $\Omega(\tau_c\log(1/\Delta_a))$ times to distinguish it from the optimal action. To see that, consider converging reward sequences that are identical for all actions for the first $\tau_c\log(1/\Delta_a)$ turns and start differing only after. This implies a worst-case lower bound of $\Omega(\tau_c\Delta_a\log(1/\Delta_a))$ for the noiseless case (Theorem \ref{lower-bound} in the appendix).

\sloppy The exponential increase in the epoch lengths is chosen to obtain bound of the form $\mathcal{O}(\tau_c\log(1/\Delta_a))$ in Theorem \ref{thm-noiseless} part (a). Any increasing sequence of epoch lengths will give a finite regret but will not have a better bound orderwise. For example, linearly increasing epoch lengths i.e., $\ell_n\propto m_{a_n,n}$, yield a bound of $\mathcal{O}((\tau_c\log(1/\Delta_a))^2)$. On the other hand, even if epoch lengths grew faster than exponential, e.g., $\ell_n\propto \exp(\exp(m_{a_n,n}))$, we would still obtain a bound of $\mathcal{O}(\tau_c\log(1/\Delta_a))$. Epoch lengths that increase too fast do not do well in practice as they waste precious time on suboptimal actions. 

\subsection{The Noisy Case}
We now consider the general case where $\eta_t\neq0$. The UECB algorithm given for the noiseless case cannot be used here as it only considers the final reward observed, which can be very noisy. To deal with the noise, it is necessary to average multiple observations. The estimated expected reward corresponding to the equilibrium point of an action has two kinds of errors - due to the i.i.d. noise and due to the distance from equilibrium (i.e., ``equilibrium noise''). Averaging creates a trade-off between the two errors. For example, averaging over all rewards observed for an action reduces the i.i.d.\ noise but increases the equilibrium noise since early rewards were earned far from equilibrium. Similarly, considering only the last reward has low equilibrium noise but high i.i.d.\ noise.

Hence we propose the UECB algorithm for the noisy case inspired by the popular UCB algorithm \cite{bandit-book}. We have the same epoch-based structure as before, i.e., $\ell_{n+1}\coloneqq 2\rho_2\exp\left(\rho_1(m_{a_{n+1},n}+1)\right)$. For action $a_{n}$, define
$$\hat{x}_{a_{n},n}=\frac{2}{\ell_n}\sum_{t=t_{n-1}+\frac{\ell_{n}}{2}+1}^{t_{n}}y_t,$$
i.e., the average of the second half of the last epoch corresponding to that action. For other actions $a\neq a_n$, $\hat{x}_{a,n}=\hat{x}_{a,n-1}$. Also, for all actions $a$, define $n_a$ as the last epoch action $a$ was played before the end of epoch $n$. Then $\ell_{n_a}$ is the length of the last epoch action $a$ was played before the end of epoch $n$, i.e., $\ell_{n_a}=2\rho_2\exp\left(\rho_1m_{a,n}\right)$. Then for all actions, we define:
$$UECB_{a,n}\coloneqq\hat{x}_{a,n}+\frac{2}{\ell_{n_a}}\frac{L\exp(-\frac{1}{\tau_c}(1+\frac{\ell_{n_a}}{2}))}{1-\exp(-\frac{1}{\tau_c})}+\sqrt{\frac{4\sigma^2}{\ell_{n_a}}\log\left(\frac{2}{\delta_n}\right)},$$
where $\delta_n\coloneqq \frac{1}{t_n^3}.$ Similar to the intuition behind the UCB algorithm, this $UECB_{a,n}$ is defined to ensure that $UECB_{a,n}\geq x^*_a$ with probability of at least $1-\delta_n$. This is proved in Lemma \ref{lemma-part1}. Finally, as before, the action for the next epoch is chosen as follows:
$$a_{n+1}=\argmax_{a\in\mathcal{A}}UECB_{a,n}.$$ 

\begin{algorithm}[t]
\caption{\label{algo-noisy} UECB Algorithm}

\textbf{Initialization: }Let $m_{a,0}=0$ for all $a\in\mathcal{A}$.

\textbf{Input: } Constant $\tau_c\geq 1, L>0$ and parameters $\rho_1, \rho_2>0$.

\textbf{For epochs $n=1$ to $K$ do}
\begin{enumerate}
    \item Play action $a_n=n$ for $\ell_n=2\rho_2e^{\rho_1}$ timesteps from $t=t_{n-1}+1$ to $t=t_n=t_{n-1}+\ell_n$.
    \item $\hat{x}_{a,n}, m_{a,n}, UECB_{a,n}=UPDATE(y_{t_{n-1}+1},\ldots,y_{t_n})$
\end{enumerate}

\textbf{For epochs $n\geq K+1$ do}
\begin{enumerate}
    \item Choose action $a_n=\argmax_{a\in\mathcal{A}} UECB_{a,n-1}$.
    \item Play action $a_n$ for $\ell_n=2\rho_2\exp(\rho_1(m_{a_n,n-1}+1))$\\ timesteps from $t=t_{n-1}+1$ to $t=t_n=t_{n-1}+\ell_n$.
    \item $\hat{x}_{a,n}, m_{a,n}, UECB_{a,n}=UPDATE(y_{t_{n-1}+1},\ldots,y_{t_n})$
\end{enumerate}
\textbf{End}

\textbf{function $UPDATE(y_{t_{n-1}+1},\ldots,y_{t_n})$}
\begin{enumerate}
    \item $m_{a_n,n}=m_{a_n,n-1}+1$ and $m_{a,n}=m_{a,n-1}$ for $a\neq a_n$.
    \item if (noiseless):
    \begin{enumerate}
        \item $\hat{x}_{a_n,n}=y_{t_n}$ and $\hat{x}_{a,n}=\hat{x}_{a,n-1}$ for $a\neq a_n$.
        \item $UECB_{a,n}=\hat{x}_{a,n}+Le^{-\frac{1}{\tau_c}\ell_{n_a}}$ for $a\in\mathcal{A}$.
    \end{enumerate}
    \item if (noisy):
    \begin{enumerate}
        \item $\hat{x}_{a_n,n}=\frac{2}{\ell_n}\sum_{t=t_{n-1}+\frac{\ell_n}{2}+1}^{t_n}y_t$ and $\hat{x}_{a,n}=\hat{x}_{a,n-1}$ for $a\neq a_n$.
        \item For all actions $a\in\mathcal{A}$, $$UECB_{a,n}=\hat{x}_{a,n}+\frac{2}{\ell_{n_a}}\frac{L\exp(-\frac{1}{\tau_c}(1+\frac{\ell_{n_a}}{2}))}{1-\exp(-\frac{1}{\tau_c})}+\sqrt{\frac{4\sigma^2}{\ell_{n_a}}\log\left(\frac{2}{\delta_n}\right)},$$ where $\delta_n=1/t_n^3$.
    \end{enumerate}
\end{enumerate}
\end{algorithm}

Theorem \ref{thm-noisy} gives a regret bound on the UECB algorithm. Next, we make a few comments on the algorithm and its bound:
\begin{itemize}
    \item The first term in the regret bound, $\mathcal{O}(\log(T)/\Delta_a)$, appears also in the regret bound for the UCB algorithm in stochastic multi-armed bandits and stems from the noisy observations that both scenarios share. The second term in the regret bound, $\mathcal{O}(\tau_c\log(\tau_c))$, also appears in the regret bound for the noiseless case given in Theorem \ref{thm-noiseless}. This second term dominates the first term if convergence to equilibrium takes significant time which is the case in applications. 
        %initially but it becomes insignificant asymptotically.
    \item \sloppy Theorem \ref{lower-bound} gives a lower bound of $\Omega(\log(T)/\Delta_a+\tau_c\Delta_a\log(1/\Delta_a))$ for equilibrium bandits. The first and second terms in this lower bound are obtained using the lower bounds for stochastic and noiseless equilibrium bandits, respectively, both of which are special cases of equilibrium bandits. Hence, UECB is order-wise optimal in $T$ and $\Delta_a$ while being optimal up to logarithmic factors in $\tau_c$.
    \item For each action, UECB only uses the rewards observed during the current epoch. It also uses only the latter half of that epoch. We chose the `half' fraction arbitrarily for simplicity and any other constant fraction yields the same order of magnitude dependencies. Another possible modification is to employ a weighted average over all the past rewards \cite{discounted_ucb}.
\end{itemize}

\section{Regret Analysis}
In this section, we explain our proof strategy by breaking the proof of Theorem \ref{thm-noisy} into lemmas. While the proof generally follows that of UCB \cite{bandit-book}, significant modifications are needed due to the converging nature of the rewards, and the epoch-based structure. In particular, UECB has to balance the tradeoff between the i.i.d. noise and the ``equilibrium noise'', which measures the distance to equilibrium. This tradeoff is unique to our problem. 

The first lemma gives a probabilistic bound on $|\hat{x}_{a,n}-x_a^*|$ and motivates the definition of $UECB_{a,n}$. This gives a probabilistic upper bound on how far away our estimate of the equilibrium reward is. The difference between $\hat{x}_{a,n}$ and $x_a^*$ has two terms - one due to the i.i.d. noise and the other due to ``equilibrium noise''. 
\begin{lemma}\label{lemma-part1}
The UECB algorithm maintains the following statements: 
\begin{enumerate}[(a)]
    \item The inequality
    $$|\hat{x}_{a,n}-x_a^*|\leq\sqrt{\frac{4\sigma^2}{\ell_{n_a}}\log\left(\frac{2}{\delta_n}\right)}+ \frac{2}{\ell_{n_a}}\frac{L\exp(-\frac{1}{\tau_c}(1+\frac{\ell_{n_a}}{2}))}{1-\exp(-\frac{1}{\tau_c})},$$
    holds with probability of at least $1-\delta_n$. 
    \item $UECB_{a,n}\geq x_a^*$ with probability of at least $1-\delta_n$.
    \item Define 
    \begin{equation*}
        \ell^{(1)}_{a,n}\coloneqq \frac{64\sigma^2}{\Delta_a^2}\log\left(\frac{2}{\delta_n}\right)  \; \textrm{and} \;\;\;\; \ell^{(2)}_{a,n}\coloneqq 2\tau_c\log\left(\frac{8L}{\Delta_a}\right).  
    \end{equation*}
    Then given that $\ell_{n_a}\geq \ell^{(1)}_{a,n}$ and $\ell_{n_a}\geq \ell^{(2)}_{a,n}$ we have $\hat{x}_{a,n}\leq x_a^*+\frac{\Delta_a}{2}$ with probability greater than $1-\delta_n$.
\end{enumerate}
\end{lemma}
Part (b) and (c) above are direct implications of part (a) and the definition of $UECB_{a,n}$. The conditions on $\ell_{n_a}$ in part (c) can be easily translated to conditions on $m_{a,n}$, i.e., the number of epochs during which $a$ has been played. Let the corresponding number of epochs be $m_{a,n}^{(1)}$ and $m_{a,n}^{(2)}$ respectively, i.e., $\ell^{(i)}_{a,n}=2\rho_2\exp\left(\rho_1m^{(i)}_{a,n}\right)$ for $i=1,2$. Here $m_{a,n}^{(1)}$ is the number of epochs required for the noise term to be sufficiently small, and $m_{a,n}^{(2)}$ is the number of epochs required for the second term, stemming from the convergence time, to be sufficiently small. A large $m_{a,n}$ implies that $a$ was played consecutively for a higher number of turns which reduces the error due to i.i.d. noise and brings the system closer to equilibrium. 

The next lemma shows that with high probability, the $UECB$ algorithm identifies suboptimal arms given that they have been played for a sufficiently large number of times. It also gives an upper bound on the probability of playing a suboptimal action given that the action has been played a sufficient number of times. 
\begin{lemma}\label{lemma-part2}
At the end of epoch $n$, if a suboptimal arm $a\neq a^*$ has been played for enough epochs, such that $\ell_{n_a}\geq \max\{\ell^{(1)}_{a,n},\ell^{(2)}_{a,n}\}$ (defined in Lemma \ref{lemma-part1}) then $UECB_{a,n}\leq UECB_{a^*,n}$ with probability at least $1-2\delta_n$. Therefore, under these conditions, the probability that the UECB algorithm plays action $a$ in the $(n+1)$\textsuperscript{th} epoch is bounded by:
$$P\left(a_{n+1}=a\;|\;\ell_{n_a}\geq \ell^{(1)}_{a,n}, \ell_{n_a}\geq \ell^{(2)}_{a,n}\right)\leq 2\delta_n.$$
\end{lemma}

Note that the expected instantaneous loss at time $t$ when action $a$ is taken can be split as follows: 
\begin{equation}\label{random-split}
x^*_{a^*}-x_t = (x^*_{a^*}-x^*_{a})+(x^*_{a}-x_t).
\end{equation}
The first term in \eqref{random-split} is the difference in the rewards at equilibrium between the optimal arm and a suboptimal arm. We first bound the regret corresponding to the first term, which depends on the number of times an action is taken multiplied by the suboptimality gap. The next lemma bounds the expected number of times \textit{(and not epochs)} a suboptimal action is played. 

\begin{lemma}\label{lemma-part3}
For any suboptimal arm $a\neq a^*$, the expected number of timesteps the UECB algorithm plays $a$ can be bounded as follows:
$$\E\left[\sum_{k=1}^n\ell_{k}I\{a_{k}=a\}\right]=\mathcal{O}\left(\frac{\sigma^2\log(t_n)}{\Delta_a^2}+\tau_c\log\left(\frac{1}{\Delta_a}\right)\right)$$
where $I\{\cdot\}$ is $1$ when $\{\cdot\}$ is true and $0$ otherwise (i.e., an indicator). 
\end{lemma}

The second term in (\ref{random-split}) denotes the convergence error, or ``equilibrium noise''. The regret accumulated due to this term can be shown to be $\mathcal{O}(\tau_c)$ times the number of arm switches, which is in turn bounded by the number of epochs where suboptimal actions were played. The next lemma bounds this number.
\begin{lemma}\label{lemma-part4}
For any suboptimal arm $a\neq a^*$, the expected number of epochs the UECB algorithm chooses $a$ can be bounded as follows:
$$\E\left[\sum_{k=1}^nI\{a_{k}=a\}\right]= \mathcal{O}\left(\log\left(\frac{\sigma^2\log(t_n)}{\Delta_a^2}\right)+\log\left(\tau_c\log\left(\frac{1}{\Delta_a}\right)\right)\right).$$
\end{lemma}
This gives rise to the final term in the bound in Theorem \ref{thm-noisy}. Combining Lemma \ref{lemma-part3} with Lemma \ref{lemma-part4} gives us the following final lemma which gives a regret bound at the end of each epoch.

\begin{lemma}\label{lemma-part5}
Let $t_n$ denote the time at the end of epoch $n$, then the UECB algorithm maintains:
\begin{eqnarray*}
\E[R(t_n)]&=& \mathcal{O}\bigg(\sum_{a\in\mathcal{A},a\neq a^*}\frac{\sigma^2\log(t_n)}{\Delta_a}+\tau_c\log\left(\tau_c\log\left(\frac{1}{\Delta_a}\right)\right)\nonumber\\
    &&\;\;\;\;\;\;+\tau_c\log\left(\frac{\sigma^2\log(t_n)}{\Delta_a^2}\right)\bigg).
\end{eqnarray*}
\end{lemma}
The lemma above proves our UECB regret bound, but only at timesteps that are at the end of some epoch. To complete the proof of Theorem \ref{thm-noisy}, we just need to prove a similar regret bound for all $T$. To that end, define $\tau$ as the time at which the last epoch ended, i.e., the ongoing epoch started at $t=\tau+1$. Then we divide $[1,T]$ into two intervals: $[1,\tau]$ and $[\tau+1,T]$. The lemma above gives a bound for the regret accumulated during $[1, \tau]$. The result for Theorem \ref{thm-noisy} is obtained by showing that the regret achieved during $[\tau+1,T]$ is bounded by a constant times the regret achieved during $[1,\tau]$.

\section{Applications}\label{sec-appli}
In this section, we detail two real-life problems that the equilibrium bandits framework can model. The agent is a policymaker who learns the policy that maximizes the collective good. However, the impact of any policy cannot be seen instantly as society interacts in a game-theoretic manner and converges to an equilibrium. 

\subsection{SIS Epidemic Model}\label{subsec-SIS}
We consider the Susceptible-Infectious-Susceptible (SIS) model of epidemics \cite{SIS1, SIS2} as the underlining unknown dynamics to be controlled by the decision-maker. In this model, a susceptible individual becomes infected with some probability after contacting an infected individual and remains infected for a random period of time. Once the individual recovers, they return to the susceptible class since the disease does not provide any long-lasting immunity. Examples for diseases that follow the SIS model are influenza, meningitis, and tuberculosis \cite{SIS3}. 

Specifically, we consider the networked SIS model with a graph with $M$ nodes and a symmetric weighted adjacency matrix $A$. These nodes can represent communities, cities, or countries. The weight $A_{ij}$ in $A$ is the contact probability between nodes $i$ and $j$. Let $I(i,t)$ be the fraction of infected individuals in node $i$ at time $t$. Then the discretized differential equation is given by:
\begin{equation}\label{sis-eqn-main}
    I(i,t+\Delta t)=I(i,t)+\left(\beta(1-I(i,t))\sum_{j=1}^M A_{ij}I(j,t)-\gamma I(i,t)\right)\Delta t.
\end{equation}
Here $0<\Delta t\ll 1$ is the stepsize for discretization which is assumed to be sufficiently small. $\beta>0$ is the infection rate and $\gamma>0$ is the recovery rate. Let $I(t)$ be the $M$-dimensional vector with the elements $I(i,t)$, then equation (\ref{sis-eqn-main}) can be written as
\begin{eqnarray}\label{sis-eqn-vector}
    I(t+\Delta t)&=&h(\beta,\gamma,A,\Delta t; I(t))\nonumber\\
    &\coloneqq& I(t)+\left(\beta(\mathbbm{1}_M-\textrm{diag}(I(t)))AI(t)-\gamma I(t)\right)\Delta t,
\end{eqnarray}
where $\mathbbm{1}_M$ is the identity matrix of dimension $M$ and $\textrm{diag}(I(t))$ is a diagonal matrix with the elements of the vector $I(t)$.

The decision-maker is the government or the policy-maker. Examples of actions can be the enforcement of masks, advertisements to increase awareness, and different types of lockdowns, e.g., shutting down schools or offices. These actions change the contact patterns between individuals and the rate of infection, which dictate the adjacency matrix and infection rate for this action,  denoted by $A_a$ and $\beta_a$, respectively, for action $a$. $\Delta t$ is the stepsize for the discretization and is typically much smaller than the time step at which the infection rates or the rewards are actually observed, which can range from a few days to a few weeks. 
Let $z_t=I(t)$ be the state of the system at time $t$. Then, for action $a$, $$z_{t+1}=g(a;z_t)=h^{(1/\Delta t)}(\beta_a, \gamma, A_a, \Delta t;z_t).$$ The cost function $-f(a;\cdot)$ can be a combination of the operation cost of the policies and the health damages due to the disease.

The government does not know the functions $f$ and $g$. In particular, it is unlikely that the government can estimate the matrix $A_a$ for each action in a large-scale setting. In addition, depending on the resolution, the government may or may not know the fraction of infected individuals in each node,  (e.g., neighborhoods as opposed to individuals). Fortunately, as explained next, UECB only requires weak bounds on the problem parameters to perform very well. 

Let $\lambda^{max}_a$ be the maximal eigenvalue of $A_a$. We assume that $\beta_a\lambda^{max}_a>\gamma_a$ for all actions. This assumption implies that a non-zero stable equilibrium point exists for the iteration given by (\ref{sis-eqn-vector}) for all actions. Let this equilibrium be $I_a^*$. The zero vector (i.e., no infections) is an unstable equilibrium in this case. If this assumption were false, i.e., $\beta_a\lambda^{max}_a\leq\gamma_a$, then there exists only one equilibrium point given by the zero vector which is also stable. We only consider actions that satisfy our assumption because, unfortunately, there are often no policies that completely eradicate the epidemic. Next, we show that this $g$ satisfies Assumption \ref{assu-evolution}):

\begin{proposition}\label{sis-prop}
For each action $a$, let $I_a^*$ be the non-zero stable equilibrium of the iteration in (\ref{sis-eqn-vector}), where $\beta_a, \gamma$ and $A_a$ satisfy $\beta_a\lambda_a^{max}>\gamma$ for the maximal eigenvalue of $A_a$, $\lambda_a^{max}$. If $I(t)>0$, then
\begin{equation}\label{sis-bound}
    \|I(t+\Delta t)-I_a^*\|_1\leq \max_{1\leq i\leq M} \left(1-\beta_a\Delta t\sum_{j=1}^MA_{a_{ij}}I(j,t)\right)\|I(t)-I_a^*\|_1,
\end{equation}
where $\|\cdot\|_1$ denotes the $\ell_1$ norm.
\end{proposition}

We just need a bound on $\beta_{a}$ and on the elements of matrix $A_{a}$ for all actions $a$ to employ the bound in \eqref{sis-bound}. If the policymaker does not know the number of infections in each node (but only the ``reward''), then they just need a bound on the number of infections. This bound is easy to deduce due to our assumption that $\beta_a\lambda_a^{max}>\gamma$. Let $\beta_a\sum_{j=1}^MA_{a_{ij}}I(j,t)>\alpha$ for all nodes $i$ and all actions. Then, 
\begin{align*}
   \|z_{t+1}-z_a^*\|_1&\leq  \left(\max_{1\leq i\leq M} \left(1-\beta_a\Delta t\sum_{j=1}^MA_{a_{ij}}I(j,t)\right)\right)^{1/\Delta t}\|z_t-z_a^*\|_1 \\
   &\lessapprox e^{-\alpha}\|z_t-z_a^*\|_1.
\end{align*}
Then $c(a;z)$ as defined in equation (\ref{assumption-2}) is approximately equal to $e^{-\alpha}$, which becomes accurate as $\Delta t$ goes to $0$. If the number of infections in each node is known, then better bounds can be obtained.

\subsection{Strongly Monotone Game}
Our second example is deterministic gradient-based learning in continuous games \cite{mazu}. The underlying system in this example is a set of $M$ players. Each player $i\in\mathcal{M}=\{1,\ldots,M\}$ has their own decision variable $z_{i,t}\in\mathcal{Z}_i\subset\mathbb{R}^d$ at time $t$. Here each $\mathcal{Z}_i$ is a convex and compact set. The state variable of the system is just a concatenation of all the decision variables, i.e., $z_t=(z_{1,t},\ldots,z_{M,t})\in\mathcal{Z}_1\times\ldots\times\mathcal{Z}_M\subset\mathbb{R}^{dM}$. Each player has their own utility function, which depends on the decision variables of all the players and is parameterized by the action taken by the agent. At time $t$, if the action taken by the agent is $a_t$, then the utility function for player $i$ is given by $u_i(a_t;z_t)$. Each player seeks to maximize their utility function and can only control their own decision variable. We assume that all players have access to the decision variables of each player and their own utility function, but not to the utility functions of other players. Then at each time $t$, each agent updates their decision variables as follows:
\begin{equation}\label{ex-2}
z_{i,t+1}=z_{i,t}+\alpha h_i(a_t;z_t),
\end{equation}
where $h_i(a;z_t)=\frac{\partial u_i(a;z_t)}{\partial z_{i,t}}$. We have considered a constant step-size $\alpha$ for simplicity; the following results can easily be generalized to a decreasing step-size. The agent observes noisy rewards based on its own reward function: $f(a_t;z_t)$. 

We make certain assumptions on the underlying game to ensure that the assumptions for our bandit problem are satisfied. To that end, define $H(a;z)=(h_1(a;z),\ldots,h_M(a;z))$, i.e., the concatenation of all gradients. Then we assume that 
$$\langle z'-z, H(a;z')-H(a;z)\rangle\leq -\lambda_a\|z'-z\|_2^2,\; \forall a\in\mathcal{A}.$$
This implies that for all actions, $\mathcal{G}_a=(\mathcal{M}, \{\mathcal{Z}_i\},\{u_i(a;\cdot)\})$ is a strongly monotone game with parameter $\lambda_a>0$. Then for each action $a$, $\mathcal{G}_a$ has a unique pure Nash equilibrium $z_a^*$ \cite{Rosen} which acts as the equilibrium corresponding to that action. For an action $a$, the Nash equilibrium $z^*$ is defined as the decision profile which satisfies: $u_i(a;z_i^*,z_{-i}^*)\geq u_i(a;z_i,z_{-i}^*)$ for all $z_i\in\mathcal{Z}_i$ and for all $i\in\mathcal{M}$. Note that $z_{-i}$ denotes the decision variables for all players except $i$. This ensures that part (a) of Assumption \ref{assu-evolution} is satisfied. We also assume that, for all $a$, $H(a,.)$ is $\beta_{a}$-Lipschitz continuous: 
$$\|H(a;z)-H(a;z')\|_2\leq\beta_a\|z-z'\|_2. $$ The next proposition shows that such games satisfy Assumption \ref{assu-evolution}:
\begin{proposition}\label{prop:game}
The iterates given by (\ref{ex-2}), with action $a_t=a$, satisfy:
$$\|z_{t+1}-z_a^*\|_2\leq \sqrt{1-2\lambda_a\alpha+\alpha^2\beta_a^2}\|z_{t}-z_a^*\|_2.$$
\end{proposition}
For a sufficiently small step-size ($\alpha\leq \frac{2\lambda_a}{\beta_a^2}$), this proposition shows that the distance of the state $z_t$ from $z_a^*$ decreases when action $a$ is taken. A sufficiently small step-size can be avoided with a sequence of decreasing step-sizes, which instead would imply that the system satisfies Assumption \ref{assu-evolution} from some $t_0$ onwards. 

As a concrete example, consider a resource allocation game \citep{Agrawal,bistritz2021online} where there are $d$ resources and each player's decision variable $z_i=(z_i^1,\ldots,z_i^d)$ denotes how much to use of each resource. The utility function of player $i$ depends on the value it assigns to each resource and the price of each resource. Here, action $a$ by the policymaker means that each player $i$ has access only to the subset of resources $\mathcal{R}_i(a)\subseteq\{1,\ldots,d\}$, so $z_i^{\ell}=0$ for $\ell\notin\mathcal{R}_i(a)$. The reward function for the agent is the sum of utilities of the players $f(a;z)=\sum_{i=1}^{M} u_i(a;z)$. 

%where $\beta_i$ are the positive weights that the agent assigns to players. 

\begin{figure*}[!h]
\centering
\begin{subfigure}{.3\textwidth}
  \centering
  \includegraphics[width=0.99\linewidth]{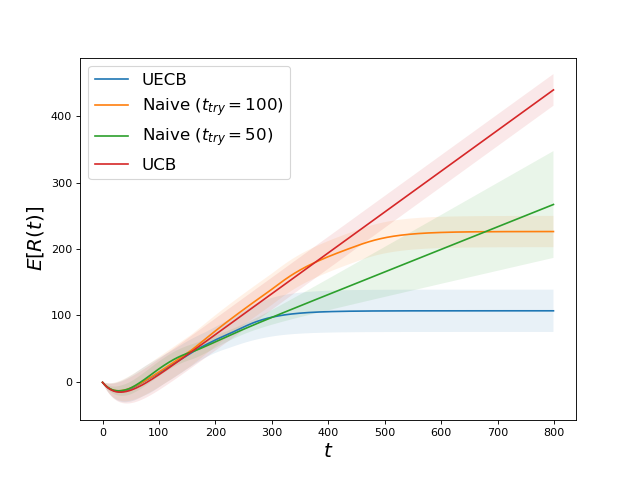}
  \caption{Noiseless}
  \label{fig:epi_noiseless}
\end{subfigure}%
\begin{subfigure}{.3\textwidth}
  \centering
  \includegraphics[width=.99\linewidth]{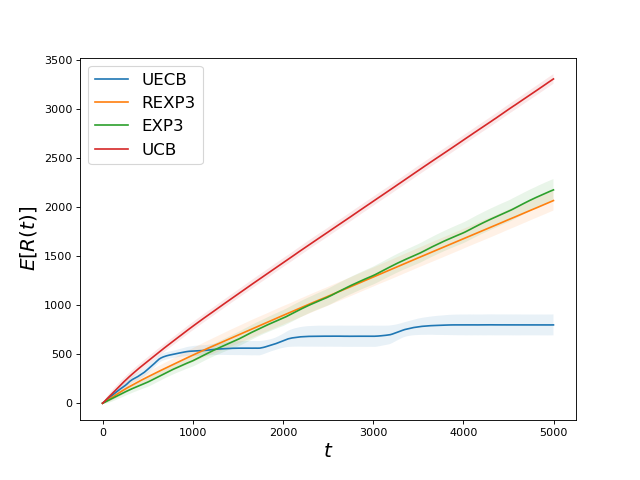}
  \caption{Noisy}
  \label{fig:epi_noisy}
\end{subfigure}
\caption{Simulation plots for SIS epidemic control in the (a) noiseless, and (b) noisy cases}
\label{fig_epi}
\end{figure*} 
\begin{figure*}[!h]
\centering
\begin{subfigure}{.3\textwidth}
  \centering
  \includegraphics[width=0.99\linewidth]{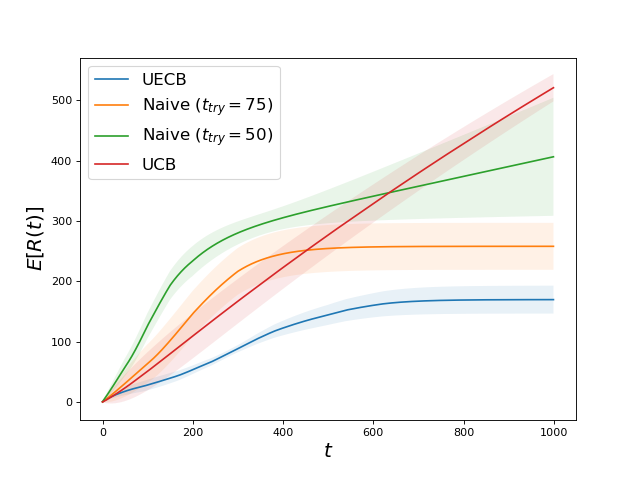}
  \caption{Noiseless}
  \label{fig:game_noiseless}
\end{subfigure}%
\begin{subfigure}{.3\textwidth}
  \centering
  \includegraphics[width=.99\linewidth]{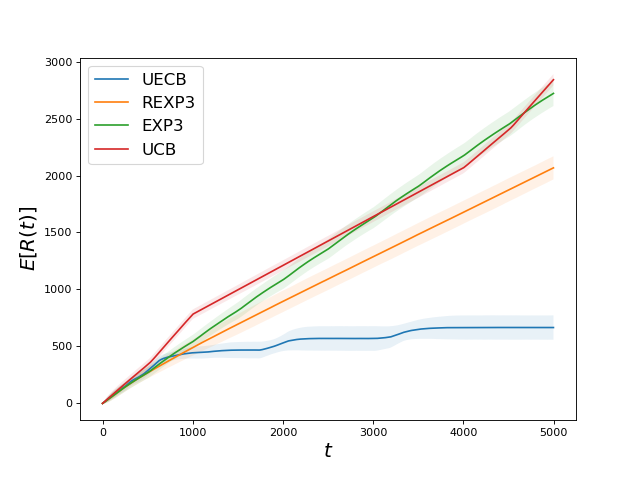}
  \caption{Noisy}
  \label{fig:game_noisy}
\end{subfigure}
\caption{Simulation plots for resource allocation game in the (a) noiseless, and (b) noisy cases.}
\label{fig_game}
\end{figure*} 

\section{Simulations}
In this section, we simulate the applications from Section \ref{sec-appli}. Each curve is the average of 100 random realizations, and has been plotted along with the standard deviation region. The randomness in the noiseless case stems from the random initializations.

\subsection{SIS Epidemic}

We simulate a system with $K=4$ actions and $M=10$ nodes. For each action $a$, we generate a random sparse symmetrical matrix $A_{a}$. We use $\gamma=0.01$  for all actions. The values for $\beta_{a}$ for the $4$ actions are $0.011,0.012, 0.013$ and $0.014$ respectively \cite{SIS3}. We use the \textbf{cost} function $f(a;z)\coloneqq w_{0,a}+w_a^Tz,$, where $w_a\in (0,1]^{M}$ is the health cost vector and $w_{0,a}\in(0,1]$ is the operational cost which only depends on the action. Clearly, $f$ is  Lipschitz with $L=1$. 

To implement UECB, we only assume that $\beta_a>0.05$ for all actions and that the sum of each row of $A$ is at least $1$, while the actual values are unknown and in $[3,5]$. Additionally, the infected fraction for each node is unknown and only the cost is known. 

Figure \ref{fig:epi_noiseless} compares the performance of UECB for the noiseless case with a naive algorithm where each action is played for $t_{try}$ consecutive timesteps in the beginning (see subsection \ref{subsec:noiseless}). Then,  the naive algorithm plays the arm that had the best reward at the end of its epoch for the rest of the timesteps. As expected, our UECB algorithm outperforms the naive algorithm for both small and large $t_{try}$. For small $t_{try}$, we do not give enough time for the system to converge which then commits to a suboptimal action, yielding linear regret. For large $t_{try}$, the system gets close to equilibrium for each action, but wastes time on suboptimal actions. This gives an $O(1)$ regret, but it is still worse than that of UECB. For the noisy case, Figure \ref{fig:epi_noisy} shows that UECB achieves sublinear regret over time while UCB, EXP3 and REXP3 \cite{besbes2014stochastic} do not.

\subsection{Resource Allocation Game}
We consider electricity grids as our resource allocation game \cite{Deng_DSM,MA_DSM}, with $d=10$ resources and $M=1000$ players. The utility function for each player $i$ is defined as
$$u_i(z)=\sum_{\ell=1}^d \left(\gamma_{i,\ell}\log(1+z_i^{\ell})-\zeta_{i,\ell}z_i^{\ell}s_{\ell}\right),$$
where $s_{\ell}=\sum_{i=1}^M z_i^{\ell}$. $\gamma_{i,\ell}$ and $\zeta_{i,\ell}$ are chosen uniformly at random in $[0.8,1]$. It can be easily verified that this function satisfies our assumptions. For each action, the subsets of resources that can be chosen by each player are generated randomly. 

We again assume very little knowledge about the system. The agent knows $\alpha$, but only uses bounds on $\lambda_a$ and $\beta_a$. Here, we assume that the agent can observe the current state, since monitoring which player picked what resource is natural in practice. The results, given in Figure \ref{fig_game}, are similar to those from the SIS epidemic scenario.

\section{Conclusions}\label{sec-conc}
In this paper, we presented equilibrium bandits, a new bandit problem, designed to deal with systems that converge to equilibrium over time. The agent can control some parameters of this system that dictate the resulting equilibrium. While the agent only observes the real-time impact of their actions, their aim is to find the set of parameters that give the best performance at equilibrium. 

We proposed Upper Equilibrium Concentration Bound (UECB), a new algorithm for equilibrium bandits that assumes very little about the system. The key innovation of UECB is the use of `convergence bounds' which bound how far the system is from the equilibrium at any given point. We proved regret bounds for UECB which are optimal up to logarithmic factors. We showed that two applications, epidemic control and resource allocation games, fall under the framework of equilibrium bandits. We simulated UECB to confirm the theoretical performance guarantees for these applications. 

By introducing a new bandit model, our work opens up many new research avenues. An important extension is to be able to learn the evolution system parameters (i.e., $\tau_c$ and $L$) on the fly, instead of using fixed worst-case bounds on these parameters. Another significant extension is to systems that evolve stochastically, which would allow the equilibrium bandits framework to include reinforcement learning algorithms and stochastic gradient-based games. 

% \bibliographystyle{ACM-Reference-Format}
% \bibliographystyle{apalike}
% \bibliography{main.bib}

\onecolumn
\newpage
  
\newpage
\appendix
\section{Appendix}
\subsection{Linear Regret of UCB Algorithm}
\begin{theorem}\label{UCB-linear}
There exist instances of equilibrium bandits satisfying Assumption \ref{assu-evolution} and with a Lipschitz continuous reward function where the UCB algorithm achieves linear regret.
\end{theorem}
\begin{proof}[\textbf{Proof}]
Consider an equilibrium bandit problem with two arms, $1$ and $2$. The equilibrium states are $z_1^*=-1$ and $z_2^*=1.5$ and the reward function is $f(a;z)=z^2$. For arm $1$, the evolution function satisfies $g(1;0.5)=-0.5$, $g(1;-0.5)=-1$ and $g(1;-1)=-1$. Similarly for arm $2$, $g(2;-0.5)=0.5$, $g(2;0.5)=1.5$ and $g(2;1.5)=1.5$. Consider noiseless rewards. At equilibrium, action $2$ is clearly optimal, i.e., $a^*=2$.  Then starting at $z_0=0.5$ at time $t=0$,  UCB would keep switching between actions $0$ and $1$ at each timestep. This would not allow either of the arms to converge, since convergence for any action requires $2$ consecutive steps. Therefore, UCB obtains linear regret in this instance.
\end{proof}

\subsection{Lower Bound for Equilibrium Bandits}
\begin{theorem}\label{lower-bound}
Consider algorithms that achieve $o(T^p)$ expected regret for all $p>0$ and all instances of equilibrium bandits (satisfying Assumption \ref{assu-evolution} and with a Lipschitz continuous reward function). Then there exist instances where all such algorithms achieve $\Omega(log(T)/\Delta_a+\tau_c\Delta_a\log(1/\Delta_a))$ regret.
\end{theorem}
\begin{proof}[\textbf{proof}]
Suppose there exists an algorithm that achieves an expected regret bound better than $O(log(T)/\Delta_a)$ for all equilibrium bandits instances. Then this algorithm would achieve regret better than $O(log(T)/\Delta_a)$ for stochastic bandits as they are a special case of equilibrium bandits where the system converges to equilibrium instantly, i.e., $g(x,a)=x_a^*$. This contradicts the known lower regret bound for stochastic bandits \cite{bandit-book}. Hence no such algorithm exists.

Next, consider a noiseless equilibrium bandit instance with converging reward sequences that are identical for the first $\tau_c\log(1/\Delta_a)$ turns and start differing only after. Specifically, suppose there are two arms, $1$ and $2$. The equilibrium states are $z_1^*=-2$ and $z_2^*=2$. For arm $1$, the evolution function is: $g(1;z)=-1$ for $z>0$ and $g(1;z)=(1-e^{-1/\tau_c})z_1^*+e^{-1/\tau_c}z$ for $z\leq 0$. For arm $2$, the evolution function is: $g(2;z)=1$ for $z<0$ and $g(2;z)=(1-e^{-1/\tau_c})z_1^*+e^{-1/\tau_c}z$ for $z\geq 0$. The reward function for arm $1$ is $f(1; z)=0$ for all $z$. For arm 2, the reward function is $f(2;z)=0$ for all $z<2-\Delta_1$ and $f(2;z)=\Delta_1+z-z_2^*$ for all $z\geq 2-\Delta_1$. Then this satisfies Assumption \ref{assu-evolution} and has a Lipschitz reward function with $L=1$. This would generate reward sequences that are identical for the first $\tau_c\log(1/\Delta_a)$ steps. If an algorithm does not differentiate between the two arms, then there exists an instance for which the algorithm incurs linear regret. To differentiate between the two arms, any algorithm has to play the suboptimal arm $1$ at least $\tau_c\log(1/\Delta_a)$ turns. Hence, an algorithm with sublinear regret has to play the suboptimal arm $1$ at least $\tau_c\log(1/\Delta_a)$ turns which incurs a regret of $\tau_c\Delta_a\log(1/\Delta_a)$. This specific instance gives us a lower bound on the regret for equilibrium bandits. 

Combining the two bounds above completes the proof.
\end{proof}

\subsection{Proof of Theorem \ref{thm-noiseless}}
\begin{proof}[\textbf{Proof of Theorem \ref{thm-noiseless}}] \textbf{(a).} Suppose a suboptimal action $a$ was taken in epoch $n$, i.e., $a_n=a$ where $a\neq a^*$. Then using equation (\ref{UECB-intuition}), we know that $ x_{t_n}-L\exp(-\ell_n/\tau_c)\leq x_a^*\leq x_{t_n}+L\exp(-\ell_n/\tau_c)$. Here $\ell_n=2\rho_2\exp(\rho_1m_{a,n}))$ is the length of the $n$\textsuperscript{th} epoch. Also, recall that $\hat{x}_{a,n}$ is defined as $x_{t_n}$ and $UECB_{a,n}=\hat{x}_{a,n}+Le^{-\ell_n/\tau_c}$. Then it directly follows that $UECB_{a,n}>x_a^*$. Now define 
$$\hat{m}_a\coloneqq\frac{1}{\rho_1}\log\left(\frac{\tau_c}{2\rho_2}\log_+\left(\frac{2L}{\Delta_a}\right)\right)+1.$$
Here and henceforth in this paper, we define $\log_+(\cdot)=\log(\max\{\cdot,1\})$ which is required when $\Delta_a$ is so large for a suboptimal arm $a$ that UECB identifies it as suboptimal in one epoch. Now suppose that $m_{a,n}=\hat{m}_a$, then  it can be verified for the corresponding $\ell_n$ that
$$Le^{-\ell_n/\tau_c}\leq \frac{\Delta_a}{2},$$
where $\Delta_a$ is the suboptimality gap for action $a$. Then 
\begin{eqnarray*}
UECB_{a,n}&=&\hat{x}_{a,n}+Le^{-\ell_n/\tau_c}\\
&\leq& x_a^* + 2Le^{-\ell_n/\tau_c} \\
&\leq& x_a^*+\Delta_a \\
&\leq& x^*_{a^*} \\
&<& UECB_{a^*,n}.
\end{eqnarray*}
This implies that action $a$ will not be taken after epoch $n$ and hence the maximum number of epochs UECB algorithm plays action $a$ is $\hat{m}_a$. Then the maximum number of timesteps a suboptimal action $a$ can be played is (denoted by $\hat{T}_a$)
\begin{eqnarray}\label{proof-noiseless-a}
\hat{T}_a&=&\sum_{m=1}^{\hat{m}_a} 2\rho_2\exp(\rho_1m_{a,n}) \nonumber\\
&\leq &\frac{e^{2\rho_1}}{e^{\rho_1}-1}\tau_c\log_+\left(\frac{2L}{\Delta_a}\right)+2\rho_2.
\end{eqnarray}
Summing $\hat{T}_a$ over all actions completes the proof of part (a) of Theorem \ref{thm-noiseless}.
\end{proof}

\begin{proof}[\textbf{Proof of Theorem \ref{thm-noiseless}}] \textbf{(b). }
At any time $t$, the instantaneous regret can be bounded as
$$x_{a^*}^*-x_t\leq |x_{a^*}^*-x_t|\leq |x_{a^*}^*-x_{a_t}^*|+|x_{a_t}^*-x_t|.$$
Then for time $T$,
\begin{equation}\label{noiseless-regret-split}
    R(T)\leq \sum_{t=1}^T |x_{a^*}^*-x_{a_t}^*| + \sum_{t=1}^T |x_{a_t}^*-x_t|.
\end{equation}
Here the first term denotes the difference in the rewards at equilibrium between the optimal and suboptimal arms. The second term denotes the regret due to the additional cost of switching between arms caused by the time taken for the system to converge to the equilibrium. The first term is simply bounded by 
$$\sum_{t=1}^T |x_{a^*}^*-x_{a_t}^*| \leq \frac{e^{2\rho_1}}{e^{\rho_1}-1}\sum_{a\in\mathcal{A},a\neq a^*}\Delta_a\tau_c\log_+\left(\frac{2L}{\Delta_a}\right)+2\rho_2\Delta_a,$$
using \eqref{proof-noiseless-a}. Now for the second term, we can split it into sums as follows, potentially completing the last epoch beyond time $T$,  $$\sum_{t} |x_{a_t}^*-x_t|\leq \sum_i \sum_{t=t^{(i)}+1}^{t=t^{(i+1)}}|x_{a_t}^*-x_t|,$$ where $t^{(i)}$ denote the times at which epochs ended and an action was switched. Then,
$$\sum_{t=t^{(i)}+1}^{t=t^{(i+1)}}|x_{a_t}^*-x_t|\leq \sum_{t=t^{(i)}+1}^{t=t^{(i+1)}} L\exp(-\frac{1}{\tau_c}(t-t^{(i)}))\leq L\frac{e^{-\frac{1}{\tau_c}}}{1-e^{-\frac{1}{\tau_c}}}\leq L\tau_c.$$
This implies that $\sum_{t} |x_{a_t}^*-x_t|\leq \sum_iL\frac{e^{-1/\tau_c}}{1-e^{-1/\tau_c}}$. We know that the total number of epochs for each suboptimal action $a$ is bounded by $\hat{m}_a$. Then the total number of switches that the algorithm $UECB$ makes is bounded by $2\sum_{a\in\mathcal{A},a\neq a^*} \hat{m}_a$. So,
$$\sum_{t} |x_{a_t}^*-x_t|\leq L\tau_c \sum_{a\in\mathcal{A},a\neq a^*}2\hat{m}_a\leq  \sum_{a\in\mathcal{A},a\neq a^*}\frac{2L}{\rho_1}\tau_c\log_+\left(\frac{\tau_c}{2\rho_2}\log_+\left(\frac{2L}{\Delta_a}\right)\right)+1.$$
Combining the bounds for the two terms in (\ref{noiseless-regret-split}) completes the proof of Theorem \ref{thm-noiseless}.
\end{proof}

\subsection{Proof of Lemma \ref{lemma-part1}}
\begin{proof}[\textbf{Proof of Lemma \ref{lemma-part1}}] \textbf{(a).} 
After the completion of epoch $n$,
$$\hat{x}_{a,n}=\frac{2}{\ell_{n_a}}\sum_{t=t'_a+\frac{\ell_{n_a}}{2}}^{t''_a}y_t,$$
where the interval $[t'_a,t''_a]$ is the last epoch in which action $a$ was played. Now,
\begin{eqnarray*}
|\hat{x}_{a,n}-x_a^*|&\leq& \frac{2}{\ell_{n_a}}\left|\sum_{t=t'_a+\frac{\ell_{n_a}}{2}}^{t''_a}\left(y_t-x_t\right)\right|+\frac{2}{\ell_{n_a}}\sum_{t=t'_a+\frac{\ell_{n_a}}{2}}^{t''_a}|x_t-x_a^*|\\
&=& \frac{2}{\ell_{n_a}}\left|\sum_{t=t'_a+\frac{\ell_{n_a}}{2}}^{t''_a}\eta_t\right|+\frac{2}{\ell_{n_a}}\sum_{t=t'_a+\frac{\ell_{n_a}}{2}}^{t''_a}|x_t-x_a^*|\\
&\stackrel{(i)}{\leq}& \frac{2}{\ell_{n_a}}\left|\sum_{t=t'_a+\frac{\ell_{n_a}}{2}}^{t''_a}\eta_t\right|+ \frac{2}{\ell_{n_a}} \sum_{i=0}^{\ell_{n_a}/2}L\exp\left(-\frac{1}{\tau_c}\left(1+\frac{\ell_{n_a}}{2}+i\right)\right)\\
&\leq& \frac{2}{\ell_{n_a}}\left|\sum_{t=t'_a+\frac{\ell_{n_a}}{2}}^{t''_a}\eta_t\right|+ \frac{2}{\ell_{n_a}}\frac{L\exp(-\frac{1}{\tau_c}(1+\frac{\ell_{n_a}}{2}))}{1-\exp(-\frac{1}{\tau_c})}.
\end{eqnarray*}
Here inequality (i) follows from equation (\ref{UECB-intuition}). Now using Chernoff bound for subgaussian random variables (Proposition 2.5 from \cite{wain_prob}), we have that
$$\frac{2}{\ell_{n_a}}\left|\sum_{t=t'_a+\frac{\ell_{n_a}}{2}}^{t''_a}\eta_t\right|\leq \sqrt{\frac{2\sigma^2}{\ell_{n_a}/2}\log\left(\frac{2}{\delta}\right)}$$
holds with probability greater than or equal to $1-\delta$. Substituting $\delta=\delta_n=1/t_n^3$ completes the proof of part (a).
\end{proof}

\begin{proof}[\textbf{Proof of Lemma \ref{lemma-part1}}] \textbf{(b).} 
Recall that 
\begin{equation*}
    UECB_{a,n}=\hat{x}_{a,n}+\frac{2}{\ell_{n_a}}\frac{L\exp(-\frac{1}{\tau_c}(1+\frac{\ell_{n_a}}{2}))}{1-\exp(-\frac{1}{\tau_c})}+\sqrt{\frac{4\sigma^2}{\ell_{n_a}}\log\left(\frac{2}{\delta_n}\right)}.
\end{equation*}

Using, Lemma \ref{lemma-part1} (a), we know that 
$$x_a^*\leq \hat{x}_{a,n}+ \frac{2}{\ell_{n_a}}\frac{L\exp(-\frac{1}{\tau_c}(1+\frac{\ell_{n_a}}{2}))}{1-\exp(-\frac{1}{\tau_c})}+\sqrt{\frac{4\sigma^2}{\ell_{n_a}}\log\left(\frac{2}{\delta_n}\right)},$$
holds with probability at least 1-$\delta_n$. So, $UECB_{a,n}\geq x_a^*$ with probability at least $1-\delta_n$.
\end{proof}

\begin{proof}[\textbf{Proof of Lemma \ref{lemma-part1}}] \textbf{(c).} 
It can be verified that $\ell_{n_a}\geq \ell_{a,n}^{(1)}$ implies that 
$$\sqrt{\frac{4\sigma^2}{\ell_{n_a}}\log\left(\frac{2}{\delta_n}\right)}\leq \frac{\Delta_a}{4},$$ and $\ell_{n_a}\geq \ell_{a,n}^{(2)}$ implies that $$\frac{2}{\ell_{n_a}}\frac{L\exp(-\frac{1}{\tau_c}(1+\frac{\ell_{n_a}}{2}))}{1-\exp(-\frac{1}{\tau_c})}\leq \frac{\Delta_a}{4}.$$
Then using Lemma \ref{lemma-part1} (a), we know that 
$$\hat{x}_{a,n}\leq x_a^*+ \frac{2}{\ell_{n_a}}\frac{L\exp(-\frac{1}{\tau_c}(1+\frac{\ell_{n_a}}{2}))}{1-\exp(-\frac{1}{\tau_c})}+\sqrt{\frac{4\sigma^2}{\ell_{n_a}}\log\left(\frac{2}{\delta_n}\right)},$$
holds with probability exceeding 1-$\delta_n$. So, given $\ell_{n_a}\geq \ell^{(1)}_{a,n}$ and $\ell_{n_a}\geq \ell^{(2)}_{a,n}$, 
    $$\hat{x}_{a,n}\leq x_a^*+\frac{\Delta_a}{2},\;w.p.\; \geq 1-\delta_n.$$
\end{proof}

\subsection{Proof of Lemma \ref{lemma-part2}}
\begin{proof}[Proof of Lemma \ref{lemma-part2}]
Suppose $\ell_{n_a}$ is greater than or equal to $\ell^{(1)}_{a,n}$ and  $\ell^{(2)}_{a,n}$. Then using union bound on parts (b) and (c) of Lemma \ref{lemma-part1}, $UECB_{a^*,n}\geq x_{a^*}^*$ and $\hat{x}_{a,n}\leq x_a^*+\Delta_a/2$ together hold with probability greater than 1-2$\delta_n$. Then given that $\ell_{n_a}$ is greater than or equal to $\ell^{(1)}_{a,n}$ and  $\ell^{(2)}_{a,n}$, the following holds with probability greater than 1-2$\delta_n$.
\begin{eqnarray*}
UECB_{a,n}&=&\hat{x}_{a,n}+\frac{2}{\ell_{n_a}}\frac{L\exp(-\frac{1}{\tau_c}(1+\frac{\ell_{n_a}}{2}))}{1-\exp(-\frac{1}{\tau_c})}+\sqrt{\frac{4\sigma^2}{\ell_{n_a}}\log\left(\frac{2}{\delta_n}\right)} \\
&\stackrel{(i)}{\leq}& \hat{x}_{a,n}+ \frac{\Delta_a}{2} \\
&\stackrel{(ii)}{\leq}& x_a^*+\Delta_a\\
&\leq & x_{a^*}^*\\
&\stackrel{(iii)}{\leq} & UECB_{a^*,n}.
\end{eqnarray*}
Inequality (i) follows from the condition that $\ell_{n_a}\geq \ell^{(j)}_{a,n}$ for $j=1,2$. Inequality (ii) follows from part (c) of Lemma \ref{lemma-part1} and inequality (iii) follows from part (b) of Lemma \ref{lemma-part2}. Since $UECB_{a,n}> UECB_{a^*,n}$ with probability less than $2\delta_n$, action $a$ is chosen in the next epoch with probability lower than $2\delta_n$, i.e.,
$$P\left(a_{n+1}=a\;|\;\ell_{n_a}\geq \ell^{(1)}_{a,n}, \ell_{n_a}\geq \ell^{(2)}_{a,n}\right)\leq 2\delta_n.$$
\end{proof}

\subsection{Proof of Lemma \ref{lemma-part3}}
\begin{proof}[Proof of Lemma \ref{lemma-part3}]
We need to bound the expected number of times a suboptimal action is played. Recall that this is given by $E\left[\sum_{k=1}^n\ell_{k}I\{a_{k}=a\}\right]$. Then note that 
\begin{align*}
&\E\left[\sum_{k=1}^n\ell_{k}I\{a_{k}=a\}\right]\\
&=2\rho_2e^{\rho_1}+\E\left[\sum_{k=K}^{n-1}\ell_{k+1}I\{a_{k+1}=a\}I\left\{\{m_{a,k}< m_{a,k}^{(1)}\}\bigcup\{m_{a,k}<m_{a,k}^{(2)}\}\right\}\right]\\
&\;\;\;\;\;\;\;\;\;\;\;\;\;\;\;\;+\E\left[\sum_{k=K}^{n-1}\ell_{k+1}I\{a_{k+1}=a\}I\left\{m_{a,k}\geq m_{a,k}^{(1)}, m_{a,k}\geq m_{a,k}^{(2)}\right\}\right]\\
&\leq 2\rho_2e^{\rho_1}+\E\left[\sum_{k=K}^{n-1}\ell_{k+1}I\{a_{k+1}=a\}I\left\{m_{a,k}< m_{a,k}^{(1)}\right\}\right]\\
&\;\;\;\;\;\;\;\;\;\;\;\;\;\;\;\;+\E\left[\sum_{k=K}^{n-1}\ell_{k+1}I\{a_{k+1}=a\}I\left\{m_{a,k}< m_{a,k}^{(2)}\right\}\right]\\
&\;\;\;\;\;\;\;\;\;\;\;\;\;\;\;\;+\E\left[\sum_{k=K}^{n-1}\ell_{k+1}I\{a_{k+1}=a\}I\left\{m_{a,k}\geq m_{a,k}^{(1)}, m_{a,k}\geq m_{a,k}^{(2)}\right\}\right].
\end{align*}
Here $2\rho_2e^{\rho_1}$ denotes the timesteps in the first epoch played for each action. We will bound each of these three terms individually. 

For the first term, let us assume that $I(a_{k+1}=a,m_{a,k}< m_{a,k}^{(1)})$ takes value $1$ for more than $m_{a,n}^{(1)}-1$ epochs. Let $\tilde{k}$ be the epoch at which this indicator is $1$ for the $(m_{a,n}^{(1)}-1)$\textsuperscript{th} time. Then arm $a$ has been pulled for $m_{a,n}^{(1)}$ times. Then for all $k>\tilde{k}$, $m_{a,k}\geq m_{a,k}^{(1)}$ and hence the indicator cannot be $1$ for the epochs $k>\tilde{m}$. This contradicts our assumption and hence the $I(a_{k+1}=a,m_{a,k}< m_{a,k}^{(1)})$ is $1$ for less than or equal to $m_{a,n}^{(1)}$ epochs. This implies that even if $m_{a,n}\geq m_{a,n}^{(1)}$, the indicator is $1$ only for the first $m_{a,n}^{(1)}$ times and $m_{a,\tilde{k}}\leq m_{a,n}^{(1)}$. So,
\begin{eqnarray*}
2\rho_2e^{\rho_1}+\E\left[\sum_{k=K}^{n-1}\ell_{k+1}I\{a_{k+1}=a\}I\left\{m_{a,k}< m_{a,k}^{(1)}\right\}\right]&\leq& \sum_{j=1}^{m_{a,n}^{(1)}}2\rho_2\exp(\rho_1j)\\
&\leq& \frac{64 e^{\rho_1}}{e^{\rho_1}-1}\frac{\sigma^2}{\Delta_a^2} \log(2/\delta_n).
\end{eqnarray*}

Similarly for the second term, note that 
\begin{eqnarray*}
\E\left[\sum_{k=K}^{n-1}\ell_{k+1}I\{a_{k+1}=a\}I\left\{m_{a,k}< m_{a,k}^{(2)}\right\}\right]<2\rho_2e^{\rho_1}+\E\left[\sum_{k=K}^{n-1}\ell_{k+1}I\{a_{k+1}=a\}I\left\{m_{a,k}< m_{a,k}^{(2)}\right\}\right]&\leq& \sum_{j=1}^{m_{a,n}^{(2)}}2\rho_2\exp(\rho_1j)\\
&\leq& \frac{2\tau_ce^{\rho_1}}{e^{\rho_1}-1}\log_+\left(\frac{8L}{\Delta_a}\right)+\rho_2.
\end{eqnarray*}

For the third term,
\begin{eqnarray*}
\E\left[\sum_{k=K}^{n-1}\ell_{k+1}I\{a_{k+1}=a\}I\left\{m_{a,k}\geq m_{a,k}^{(1)}, m_{a,k}\geq m_{a,k}^{(2)}\right\}\right]&=& \sum_{k=K}^{n-1}\ell_{k+1}P\left(a_{k+1}=a|m_{a,k}\geq m_{a,k}^{(1)}, m_{a,k}\geq m_{a,k}^{(2)}\right)P\left(m_{a,k}\geq m_{a,k}^{(1)}, m_{a,k}\geq m_{a,k}^{(2)}\right)\\
&\leq& \sum_{k=K}^{n-1}\ell_{k+1} P\left(a_{k+1}=a|m_{a,k}\geq m_{a,k}^{(1)}, m_{a,k}\geq m_{a,k}^{(2)}\right)\\
&\leq& \sum_{k=K}^{n-1}\ell_{k+1}2\delta_k.
\end{eqnarray*}
Here the last inequality is obtained using Lemma \ref{lemma-part2}. Note that $\ell_{k+1}=\ell_{k_a}\times e^{\rho_1}\leq e^{\rho_1}t_k$. This implies that 
$$\sum_{k=K}^{n-1}\ell_{k+1}2\delta_k\leq 2e^{\rho_1}\sum_{k=1}^nt_k\frac{1}{t_k^3}\leq 2e^{\rho_1}\sum_{k=1}^n\frac{1}{k^2}.$$
So, 
$$\E\left[\sum_{k=1}^n\ell_{k}I\{a_{k}=a\}\right]\leq\frac{64\sigma^2\log(t_n)}{(e^{\rho_1}-1)\Delta_a^2}+\frac{2\tau_c}{e^{\rho_1}-1}\log_+\left(\frac{8L}{\Delta_a}\right)+4e^{\rho_1}+\rho_2.$$
\end{proof}

\subsection{Proof of Lemma \ref{lemma-part4}}
\begin{proof}[Proof of Lemma \ref{lemma-part4}]
We wish to bound the expected number of epochs a suboptimal action is chosen. Similar to proof of Lemma \ref{lemma-part3}, we split it as follows:
\begin{align*}
&\E\left[\sum_{k=1}^nI\{a_{k}=a\}\right]\\
&=1+\E\left[\sum_{k=K}^{n-1}I\{a_{k+1}=a\}I\left\{\{m_{a,k}< m_{a,k}^{(1)}\}\bigcup\{m_{a,k}<m_{a,k}^{(2)}\}\right\}\right]\\
&\;\;\;\;\;\;\;+\E\left[\sum_{k=K}^{n-1}\ell_{k+1}I\{a_{k+1}=a\}I\left\{m_{a,k}\geq m_{a,k}^{(1)}, m_{a,k}\geq m_{a,k}^{(2)}\right\}\right]\\
&\leq 1+\E\left[\sum_{k=K}^{n-1}I\{a_{k+1}=a\}I\left\{m_{a,k}< m_{a,k}^{(1)}\right\}\right]\\
&\;\;\;\;\;\;\;+\E\left[\sum_{k=K}^{n-1}I\{a_{k+1}=a\}I\left\{m_{a,k}< m_{a,k}^{(2)}\right\}\right]\\
&\;\;\;\;\;\;\;+\E\left[\sum_{k=K}^{n-1}I\{a_{k+1}=a\}I\left\{m_{a,k}\geq m_{a,k}^{(1)}, m_{a,k}\geq m_{a,k}^{(2)}\right\}\right].
\end{align*}
The term $1$ in the right-hand side above is to count for the first epoch played for each action. We will bound each of these three terms individually. 

Again, similar to proof of Lemma \ref{lemma-part3}, we have that the number of times the indicator $I(a_{k+1}=a,m_{a,k}< m_{a,k}^{(1)})$ takes value $1$ is bounded by $m_{a,n}^{(1)}-1$. So,
\begin{eqnarray*}
1+\E\left[\sum_{k=K}^{n-1}I\{a_{k+1}=a\}I\left\{m_{a,k}< m_{a,k}^{(1)}\right\}\right]&\leq& m_{a,n}^{(1)}\\
&\leq& \frac{1}{\rho_1}\log_+\left(\frac{32\sigma^2}{\rho_2\Delta_a^2}\log\left(\frac{2}{\delta_n}\right)\right).
\end{eqnarray*}
Similarly, 
\begin{eqnarray*}
\E\left[\sum_{k=K}^{n-1}I\{a_{k+1}=a\}I\left\{m_{a,k}< m_{a,k}^{(2)}\right\}\right]\leq 1+\E\left[\sum_{k=K}^{n-1}I\{a_{k+1}=a\}I\left\{m_{a,k}< m_{a,k}^{(2)}\right\}\right]&\leq& m_{a,n}^{(2)}\\
&\leq& \frac{1}{\rho_1}\log_+\left(\frac{\tau_c}{\rho_2}\log_+(\frac{8L}{\Delta_a})\right).
\end{eqnarray*}
Now, 
\begin{eqnarray*}
\E\left[\sum_{k=K}^{n-1}I\{a_{k+1}=a\}I\left\{m_{a,k}\geq m_{a,k}^{(1)}, m_{a,k}\geq m_{a,k}^{(2)}\right\}\right]&=&\sum_{k=K}^{n-1}P\left(a_{k+1}=a|m_{a,k}\geq m_{a,k}^{(1)}, m_{a,k}\geq m_{a,k}^{(2)}\right)P\left(m_{a,k}\geq m_{a,k}^{(1)}, m_{a,k}\geq m_{a,k}^{(2)}\right)\\
&\leq& \sum_{k=K}^{n-1} P\left(a_{k+1}=a|m_{a,k}\geq m_{a,k}^{(1)}, m_{a,k}\geq m_{a,k}^{(2)}\right)\\
&\leq& \sum_{k=K}^{n-1}2\delta_k \leq \sum_{k=1}^\infty \frac{2}{t_k^3}\leq \sum_{k=1}^\infty \frac{2}{k^3} \leq 4.
\end{eqnarray*}
So,
$$\E\left[\sum_{k=1}^nI\{a_{k}=a\}\right]\leq \frac{1}{\rho_1}\log_+\left(\frac{32\sigma^2}{\rho_2\Delta_a^2}\log\left(\frac{2}{\delta_n}\right)\right)+\frac{1}{\rho_1}\log_+\left(\frac{\tau_c}{\rho_2}\log_+\left(\frac{8L}{\Delta_a}\right)\right)+4.$$
This completes the proof of Lemma \ref{lemma-part4}.
\end{proof}

\subsection{Proof of Lemma \ref{lemma-part5}}
\begin{proof}[Proof of Lemma \ref{lemma-part5}]
As noted before, the expected cumulative regret at the end of epoch $n$ can be split as follows:
$$\E[R(t_n)]\leq \E\left[\sum_{t=1}^{t_n}(x_{a^*}-x_{a_t}^*)\right]+\E\left[\sum_{t=1}^{t_n}(x_{a_t}^*-y_t)\right]=\E\left[\sum_{t=1}^{t_n}(x_{a^*}-x_{a_t}^*)\right]+\E\left[\sum_{t=1}^{t_n}(x_{a_t}^*-x_t)\right].
$$
The first term can simply be written as
$$\E\left[\sum_{t=1}^{t_n}x_{a^*}-x_{a_t}^*\right]\leq \sum_{a\neq a^*}\E\left[\sum_{k=1}^{n} \ell_k I\{a_k=a\}\right]\Delta_a,$$ where we have a bound due to Lemma 3.

Now for the second term note that 
$$\sum_{t=1}^{t_n}(x_{a_t}^*-x_t)\leq \sum_i\sum_{t=t^{(i)}}^{t=t^{(i+1)}}(x_{a_t}^*-x_t),$$
where $t^{(i)}$ denotes the times at which epochs end and actions are switched. As shown in proof of Theorem \ref{thm-noiseless} (b), 
$$\sum_{t=t^{(i)}+1}^{t=t^{(i+1)}}|x_{a_t}^*-x_t|\leq L\frac{e^{-1/\tau_c}}{1-e^{-1/\tau_c}}\leq L\tau_c.$$
This implies that 
$$\sum_{t=1}^{t_n}(x_{a_t}^*-x_t)\leq \sum_i L\tau_c\leq L\tau_c \tilde{N}_n,$$
where $\tilde{N}_n$ denotes the number of times arms are switched till epoch $n$. As argued in the proof of Theorem \ref{thm-noiseless} (b), $\tilde{N}_n$ is bounded by
$$\tilde{N}_n\leq 2\sum_{a\neq a^*} \sum_{k=1}^n I\{a_k=a\},$$
that is twice the total number of times suboptimal actions are played. Hence,
$$\E\left[\sum_{t=1}^{t_n}(x_{a_t}^*-y_t)\right]\leq \E\left[L\tau_c \tilde{N}_n\right]\leq 2L\tau_c\sum_{a\neq a^*}\E\left[\sum_{k=1}^n I\{a_k=a\}\right].$$
Using the bound in Lemma \ref{lemma-part4}, we have for any $n$
$$\E[R(t_n)]\leq \sum_{a\neq a^*}\left(\frac{64\sigma^2\log(t_n)}{(e^{\rho_1}-1)\Delta_a}+\frac{2\tau_c\Delta_a}{e^{\rho_1}-1}\log_+\left(\frac{8L}{\Delta_a}\right)+(4e^{\rho_1}+\rho_2)\Delta_a+ \frac{2L\tau_c}{\rho_1}\log_+\left(\frac{32\sigma^2}{\rho_2\Delta_a^2}\log\left(\frac{2}{\delta_n}\right)\right)+\frac{2L\tau_c}{\rho_1}\log_+\left(\frac{\tau_c}{\rho_2}\log_+\left(\frac{8L}{\Delta_a}\right)\right)+8L\tau_c\right)$$
\end{proof}

\subsection{Proof of Theorem \ref{thm-noisy}}
\begin{proof}[Proof of Theorem \ref{thm-noisy}]
Using Lemma \ref{lemma-part5}, we now have a bound on the expected cumulative regret till the end of each epoch. We now wish to extend this for all times. Let $T$ be any arbitrary time and let $\tilde{n}(T)$ denote the last epoch which completed before time $T$, i.e., $t_{\tilde{n}(T)}<T$ and $t_{\tilde{n}(T)+1}>T$ (we ignore the case where $t_{\tilde{n}(T)}=T$ as we already have a bound for that). Then
\begin{eqnarray*}
E[R(T)]&=&E\left[\sum_{t=1}^T x_{a^*}^*-x_{a_t}^* + \sum_{t=1}^T x_{a_t^*} -y_t\right].
\end{eqnarray*}
For the second term, we have (as in the proof of Lemma \ref{lemma-part5}) 
\begin{eqnarray*}
    E\left[\sum_{t=1}^T x_{a_t^*} -y_t\right] &=&E\left[ \sum_{t=1}^T x_{a_t^*} -x_t\right]\\
    &\leq & L\tau_c \tilde{N}_T. 
\end{eqnarray*}
Recall that $\tilde{N}_n$ denotes the number of switches till the end of epoch $n$. With some abuse of notation, we use $\tilde{N}_T$ to denote the number of switches till timestep $T$. Then, $\tilde{N}_T\leq \tilde{N}_{\tilde{n}(T)}+1$ (there can be at most $1$ more switch). Now note that by splitting the first sum into two partitions, the sum from $t=1$ to $t=t_{\tilde{n}(T)}$ and the sum from $t=t_{\tilde{n}(T)+1}$ to $t=T$, we get
\begin{eqnarray*}
    \sum_{t=1}^T x_{a^*}^*-x_{a_t}^*\leq \sum_{k=1}^{\tilde{n}(T)}\sum_{a\neq a^*} \ell_k I\{a_k=a\}\Delta_a + \sum_{t=\tilde{n}(T)+1}^{T}\sum_{a\neq a^*} I\{a_{\tilde{n}(T)+1}=a\}\Delta_a
\end{eqnarray*}
Now,
\begin{eqnarray*}
\sum_{t=\tilde{n}(T)+1}^{T}\sum_{a\neq a^*} I\{a_{\tilde{n}(T)+1}=a\}\Delta_a&\leq& \sum_{a\neq a^*} \ell_{\tilde{n}(T)+1}I\{a_{\tilde{n}(T)+1}=a\}\Delta_a\\
&=& \sum_{a\neq a^*} e^{\rho_1} \ell_{{\tilde{n}(T)}_a}I\{a_{\tilde{n}(T)+1}=a\}\Delta_a.
\end{eqnarray*}
Now note that for any suboptimal $a\neq a^*$,
\begin{eqnarray*}
\ell_{{\tilde{n}(T)}_a}I\{a_{\tilde{n}(T)+1}=a\}\Delta_a&\leq&t_{a,\tilde{n}(T)} \Delta_a\\
&= & \sum_{k=1}^{\tilde{n}(T)}\sum_{a\neq a^*} \ell_k I\{a_k=a\}\Delta_a.
\end{eqnarray*}
This gives us
\begin{eqnarray*}
E[R(T)]\leq E\left[\left(1+e^{\rho_1}\right) \sum_{a\neq a^*}\sum_{k=1}^{\tilde{n}(T)} \ell_k I\{a_k=a\}\Delta_a + 2L\tau_c(\tilde{N}_{\tilde{n}(T)}+1)\right].
\end{eqnarray*}
Finally this implies
\begin{eqnarray*}
E[R(T)]&\leq& \left(1+e^{\rho_1}\right)\E\left[\sum_{k=1}^{\tilde{n}(T)}\sum_{a\neq a^*} \ell_k I\{a_k=a\}\Delta_a\right]+2L\tau_c(\E[\tilde{N}_{\tilde{n}(T)}]+1)\\
&\leq& \left(1+e^{\rho_1}\right) \sum_{a\neq a^*}\frac{64\sigma^2\log(t_{\tilde{n}(T)})}{(e^{\rho_1}-1)\Delta_a}+\frac{2\tau_c\Delta_a}{e^{\rho_1}-1}\log_+\left(\frac{8L}{\Delta_a}\right)+(4e^{\rho_1}+\rho_2)\Delta_a\\
&&\;\;\;\;\;\;\;\;\;\;\;+ \sum_{a\neq a^*} \frac{2L\tau_c}{\rho_1}\log_+\left(\frac{32\sigma^2}{\rho_2\Delta_a^2}\log\left(2t_{\tilde{n}(T)}^3\right)\right)+\frac{2L\tau_c}{\rho_1}\log_+\left(\frac{\tau_c}{\rho_2}\log_+\left(\frac{8L}{\Delta_a}\right)\right)+10L\tau_c\\
&\leq & \left(1+e^{\rho_1}\right) \sum_{a\neq a^*}\frac{64\sigma^2\log(T)}{(e^{\rho_1}-1)\Delta_a}+\frac{2\tau_c\Delta_a}{e^{\rho_1}-1}\log_+\left(\frac{8L}{\Delta_a}\right)+(4e^{\rho_1}+\rho_2)\Delta_a\\
&&\;\;\;\;\;\;\;\;\;\;\;+ \sum_{a\neq a^*}\frac{2L\tau_c}{\rho_1}\log_+\left(\frac{32\sigma^2}{\rho_2\Delta_a^2}\log\left(T^3\right)\right)+\frac{2L\tau_c}{\rho_1}\log_+\left(\frac{\tau_c}{\rho_2}\log_+\left(\frac{8L}{\Delta_a}\right)\right)+10L\tau_c.
\end{eqnarray*}
\end{proof}

\subsection{Proof of Proposition \ref{sis-prop}}
\begin{proof}[Proof of Proposition \ref{sis-prop}]
We drop the subscript $a$ for simplicity. Let $I^*(i)$ denote the $i$\textsuperscript{th} element of $I^*$. Then note that
\begin{eqnarray*}
I(i,t+\Delta t)-I^*(i)&\stackrel{(a)}{=}&\big(1-\gammadt\big)\big(I(i,t)-I^*(i)\big) + \Delta t\left(\beta(1-I(i,t))\sum_{j=1}^MA_{ij}I(j,t)-\beta(1-I^*(i))\sum_{j=1}^MA_{ij}I^*(j)\right) \\
&\stackrel{(b)}{=}& \left(1-\gammadt-\betadt\sum_{j=1}^MA_{ij}I^*(j)\right)\left(I(i,t)-I^*(i)\right)+\betadt(1-I(i,t))\sum_{j=1}^MA_{ij}\left(I(j,t)-I^*(j)\right) \\
&\stackrel{(c)}{=}& \left(1-\gammadt-\gammadt\frac{I^*(i)}{1-I^*(i)}\right)\left(I(i,t)-I^*(i)\right)+\betadt(1-I(i,t))\sum_{j=1}^MA_{ij}\left(I(j,t)-I^*(j)\right)\\
&=& \left(1-\gammadt\frac{1}{1-I^*(i)}\right)\left(I(i,t)-I^*(i)\right)+\betadt(1-I(i,t))\sum_{j=1}^MA_{ij}\left(I(j,t)-I^*(j)\right)\\
&\stackrel{(d)}{=}& \left(1-\betadt\sum_{j=1}^MA_{ij}\frac{I^*(j)}{I^*(i)}\right)\left(I(i,t)-I^*(i)\right)+\betadt(1-I(i,t))\sum_{j=1}^MA_{ij}\left(I(j,t)-I^*(j)\right).
\end{eqnarray*}
Here equality $(b)$ is obtained by adding and subtracting terms to simplify the expression. Equalities $(a)$, $(c)$ and $(d)$ follow from the fact that the equilibrium point is non-zero and satisfies the following equation:
$$\beta(1-I^*(i))\sum_{j=1}^MA_{ij}I^*(j)=\gamma I^*(i).$$

Under the assumption that $\Delta t$ is sufficiently small, we have 
$$|I(i,t+\Delta t)-I^*(i)|\leq\left(1-\betadt\sum_{j=1}^MA_{ij}\frac{I^*(j)}{I^*(i)}\right)\left|I(i,t)-I^*(i)\right|+\betadt(1-I(i,t))\sum_{j=1}^MA_{ij}\left|I(j,t)-I^*(j)\right|.$$
Summing this over all $1\leq i\leq M$, we get
\begin{eqnarray*}
\sum_{i=1}^M|I(i,t+\Delta t)-I^*(i)|\leq \sum_{i=1}^M\left(1-\betadt\sum_{j=1}^MA_{ij}\frac{I^*(j)}{I^*(i)}\right)\left|I(i,t)-I^*(i)\right| + \betadt\sum_{i=1}^M\sum_{j=1}^MA_{ij}(1-I(i,t))\left|I(j,t)-I^*(j)\right|.
\end{eqnarray*}
Combining similar terms and using the fact that $A_{ij}=A_{ji}, \forall i,j$ we get
\begin{eqnarray*}
\sum_{i=1}^M|I(i,t+\Delta t)-I^*(i)|&\leq& \sum_{i=1}^M\left(1-\betadt\sum_{j=1}^MA_{ij}\frac{I^*(j)}{I^*(i)}+\betadt\sum_{j=1}^MA_{ij}\left(1-I(j,t)\right)\right)\left|I(i,t)-I^*(i)\right|\\
&=& \sum_{i=1}^M\left(1-\betadt\sum_{j=1}^MA_{ij}I(j,t)+\betadt\sum_{j=1}^MA_{ij}\left(1-\frac{I^*(j)}{I^*(i)}\right)\right)\left|I(i,t)-I^*(i)\right|\\
&=& \sum_{i=1}^M\left(1-\betadt\sum_{j=1}^MA_{ij}I(j,t)\right)\left|I(i,t)-I^*(i)\right| + \betadt\sum_{i=1}^M\sum_{j=1}^MA_{ij}\left(1-\frac{I^*(j)}{I^*(i)}\right)|I(i,t)-I^*(i)| 
\end{eqnarray*}
Let $\|I\|_\infty$ denote the max norm of vector $I$, i.e., $\|I\|_\infty=\max_{i} I(i)$. For simplicity, with a little abuse of notation, define $$k_t= \max_{1\leq i\leq M} \left(1-\betadt\sum_{j=1}^MA_{ij}I(j,t)\right).$$ Then
\begin{eqnarray*}
\|I(t+\Delta t)-I^*\|_1&\leq&  \sum_{i=1}^M k_t\left|I(i,t)-I^*(i)\right| + \betadt\sum_{i=1}^M\sum_{j=1}^MA_{ij}\left(1-\frac{I^*(j)}{I^*(i)}\right)\|I(t)-I^*\|_\infty \\
&=&  k_t\|I(t)-I^*\|_1+ \betadt\sum_{i=1}^M\sum_{j>i}^NA_{ij}\left(2-\frac{I^*(j)}{I^*(i)}-\frac{I^*(i)}{I^*(j)}\right)\|I(t)-I^*\|_\infty 
\end{eqnarray*}
Note that $2-\frac{I^*(j)}{I^*(i)}-\frac{I^*(i)}{I^*(j)}\leq 0$ for all $1\leq i,j\leq M$. So,
\begin{eqnarray*}
\|I(t+\Delta t)-I^*\|_1\leq  \max_{1\leq i\leq M} \left(1-\betadt\sum_{j=1}^MA_{ij}I(j,t)\right)\|I(t)-I^*\|_1.
\end{eqnarray*}
This completes the proof of Proposition \ref{sis-prop}.
\end{proof}
\subsection{Proof of Proposition \ref{prop:game}}
\begin{proof}[Proof of Proposition \ref{prop:game}]
Recall that if $a_t=a$, then the iteration for gradient-based learning is given by:
$$z_{t+1}=z_t+\alpha H(a;z_t).$$
Then note that 
\begin{eqnarray}\label{game-proof-exp}
\|z_{t+1}-z^*_a\|^2_2&=&\|z_t-z_a^*+\alpha H(a;z_t)\|^2_2\nonumber\\
&=& \|z_t-z_a^*\|_2+2\alpha\langle H(a;z_t),z_t-z^*_a\rangle +\alpha^2\|H(a;z_t)\|^2_2\nonumber\\
&=& \|z_t-z_a^*\|_2+2\alpha\langle H(a;z_t)-H(a;z^*_a),z_t-z^*_a\rangle +\alpha^2\|H(a;z_t)\|^2_2.
\end{eqnarray}
Here the last equality follows from the fact that $z_a^*$ is a Nash equilibrium, which implies that $H(a;z_a^*)=0$. Then using the strongly monotone assumption, we have
$$\langle H(a;z_t)-H(a;z_a^*),z_t-z^*_a\rangle \leq -\lambda_a \|z_t-z_a^*\|_2^2$$
Now, we know that 
$$
\|H(a;z_t)\|_2^2=\|H(a;z_t)-H(a;z^*_a)\|^2_2\leq \beta_a^2\|z_t-z^*\|^2_2
$$
Combining these with (\ref{game-proof-exp}), we have
$$\|z_{t+1}-z^*_a\|^2_2\leq \left(1-2\alpha\lambda_a+\alpha^2\beta_a^2\right)\|z_t-z^*\|_2^2,$$
which completes the proof of Proposition \ref{prop:game}.
\end{proof}

\end{document}